\documentclass[lettersize,journal]{IEEEtran}

\usepackage[mathletters]{ucs}
\usepackage[utf8x]{inputenc}
\usepackage{epsfig}
\usepackage{amsmath}
\usepackage{bm}
\usepackage{amssymb}
\usepackage{mathabx}

\usepackage{mathrsfs}
\usepackage{caption}
\usepackage{pifont}
\usepackage{wasysym}
\usepackage{array}
\usepackage{caption}
\usepackage{color}
\usepackage{xcolor}
\usepackage{colortbl}
\usepackage{hyperref}
\usepackage{wrapfig}
\usepackage{multicol}
\usepackage{hhline}
\usepackage{booktabs} % To thicken table lines
\usepackage{multirow}
\usepackage{color}
\usepackage{xcolor}
\usepackage{colortbl}
\usepackage{wrapfig}
\usepackage{graphicx}
\newcommand{\eg}{\textit{e}.\textit{g}.}

\usepackage{multirow}
\usepackage{makecell, multirow, tabularx}
\usepackage{adjustbox}
\usepackage{tabularx}
\usepackage{xspace}
\usepackage{booktabs}
\usepackage{xurl}
\usepackage{subcaption}
\usepackage{graphicx}
\usepackage{algorithm}
\usepackage{algorithmic}
\usepackage[english]{babel}
\newtheorem{theorem}{Theorem}

\definecolor{hollywoodcerise}{rgb}{0.96, 0.0, 0.63}
\definecolor{lasallegreen}{rgb}{0.03, 0.47, 0.19}
\definecolor{hanpurple}{rgb}{0.32, 0.09, 0.98}
\definecolor{green(pigment)}{rgb}{0.0, 0.65, 0.31}

\usepackage{hyperref}
\hypersetup{colorlinks,linkcolor={red},citecolor={hollywoodcerise},urlcolor={red}}  

\begin{document}

\title{A Unified Framework for Unsupervised Domain Adaptation based on Instance Weighting}

\author{Jinjing Zhu,~\IEEEmembership{Student Member,~IEEE},  Feiyang Ye, Qiao Xiao, Pengxin Guo, Yu Zhang,~\IEEEmembership{Member, IEEE} and Qiang Yang,~\IEEEmembership{Fellow, IEEE} % <-this % stops a space
\IEEEcompsocitemizethanks{
\IEEEcompsocthanksitem J. Zhu is with Department of Computer Science and Engineering, Southern University of Science and Technology. F. Ye is with Department of Computer Science and Engineering, Southern University of Science and Technology and School of Computer Science, University of Technology Sydney. Q. Xiao is with Department of Mathematics and Computer Science,
Eindhoven University of Technology. P. Guo is with Department of Computer Science and Engineering, Southern University of Science and Technology. Y. Zhang is with Department of Computer Science and Engineering, Southern University of Science and Technology and Peng Cheng Laboratory. Q. Yang is with Department of Computer Science and Engineering, Hong Kong University of Science and Technology.\protect\\ E-mail: jinjingzhu.mail@gmail.com, yefeiyang123@live.com, qiaoxiao7282@gmail.com, 12032913@mail.sustech.edu.cn, yu.zhang.ust@gmail.com, qyang@cse.ust.hk
%\{qiaoxiao7282,yu.zhang.ust\}@gmail.com.
%yu.zhang.ust@gmail.com.
% \IEEEcompsocthanksitem Most parts of this work were done when the first author worked as a research assistant at Department of Computer Science and Engineering, Southern University of Science and Technology.
%\IEEEcompsocthanksitem $^*$Equal  contribution%\protect\\
\IEEEcompsocthanksitem This work was done when the first author worked as a research assistant at Southern
University of Science and Technology.
\IEEEcompsocthanksitem Corresponding author: Yu Zhang (yu.zhang.ust@gmail.com)
}% <-this % stops an unwanted space
}

\markboth{}%
{}


\maketitle

\begin{abstract}
Despite the progress made in domain adaptation, solving Unsupervised Domain Adaptation (UDA) problems with a general method under complex conditions caused by label shifts between domains remains a formidable task. In this work, we comprehensively investigate four distinct UDA settings including closed set domain adaptation, partial domain adaptation, open set domain adaptation, and universal domain adaptation, where shared common classes between source and target domains coexist alongside domain-specific private classes. The prominent challenges inherent in diverse UDA settings center around the discrimination of common/private classes and the precise measurement of domain discrepancy. To surmount these challenges effectively, we propose a novel yet effective method called \textbf{L}earning \textbf{I}nstance \textbf{W}eighting for \textbf{U}nsupervised \textbf{D}omain \textbf{A}daptation (\textbf{LIWUDA}), which caters to various UDA settings. Specifically, the proposed LIWUDA method constructs a weight network to assign weights to each instance based on its probability of belonging to common classes, and designs Weighted Optimal Transport (\textbf{WOT}) for domain alignment by leveraging instance weights. Additionally, the proposed LIWUDA method devises a Separate and Align (\textbf{SA}) loss to separate instances with low similarities and align instances with high similarities. To guide the learning of the weight network, Intra-domain Optimal Transport (\textbf{IOT}) is proposed to enforce the weights of instances in common classes to follow a uniform distribution. Through the integration of those three components, the proposed LIWUDA method demonstrates its capability to address all four UDA settings in a unified manner. Experimental evaluations conducted on three benchmark datasets substantiate the effectiveness of the proposed LIWUDA method.
\end{abstract}


\IEEEpubidadjcol

\section{Introduction}

\begin{figure*}[h]
 \centering
\subcaptionbox{Closed Set Domain Adaptation\label{fig:csda_illustration}}{
\includegraphics[width=0.23 \textwidth]{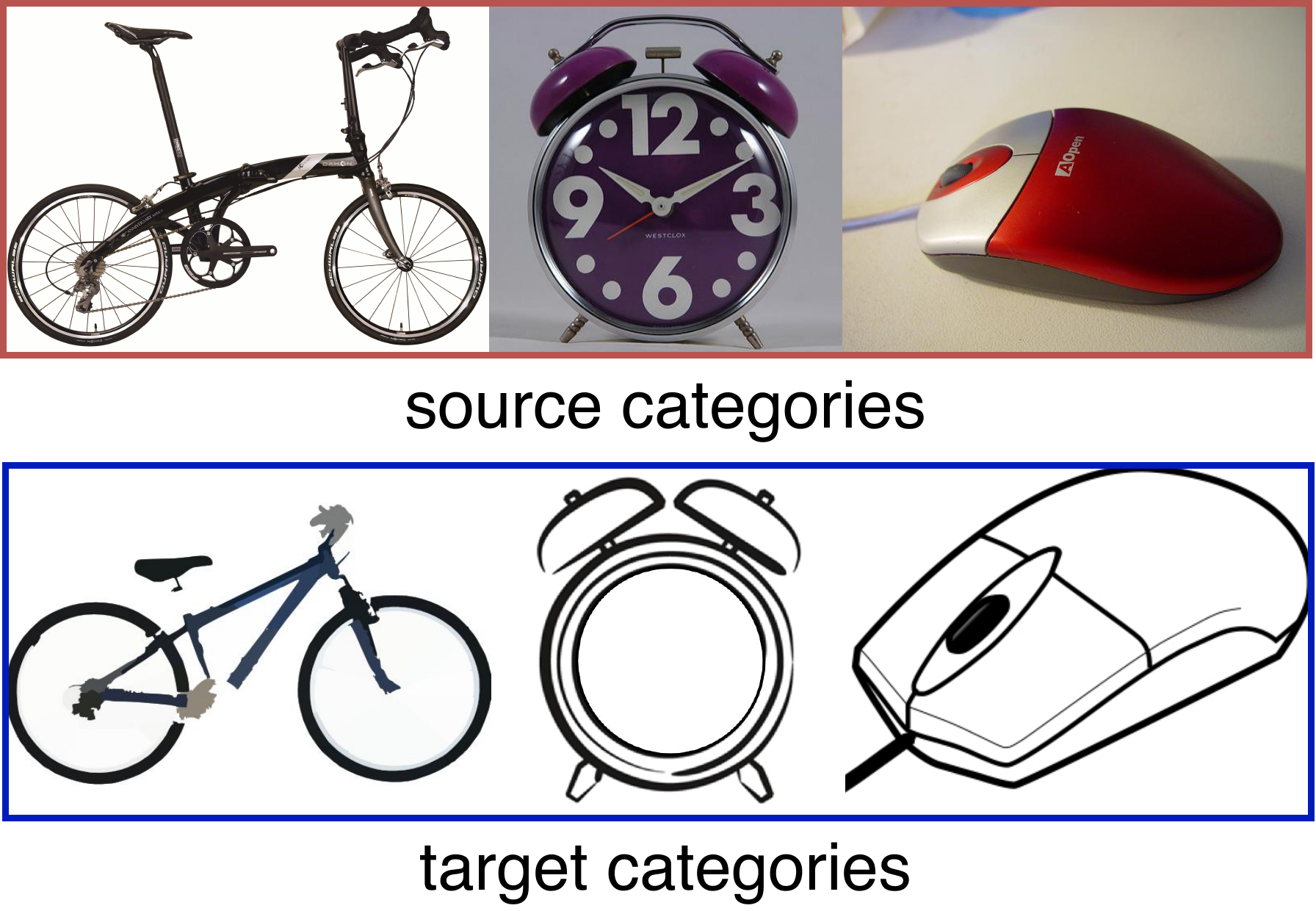}
}
\hfill
\subcaptionbox{Partial Domain Adaptation}{
\includegraphics[width=0.23 \textwidth]{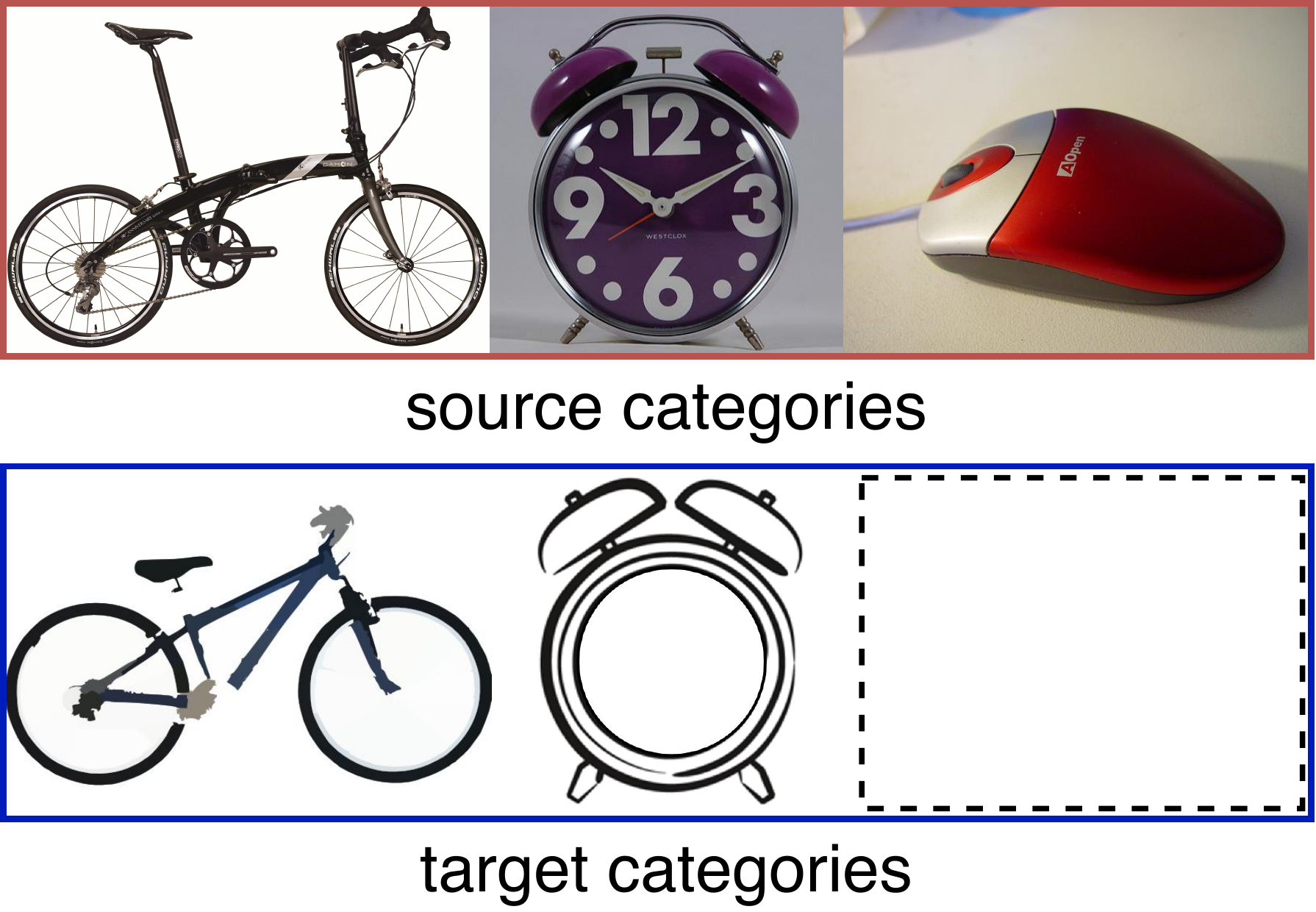}
}
\hfill
\subcaptionbox{Open Set Domain Adaptation}{
\includegraphics[width=0.23 \textwidth]{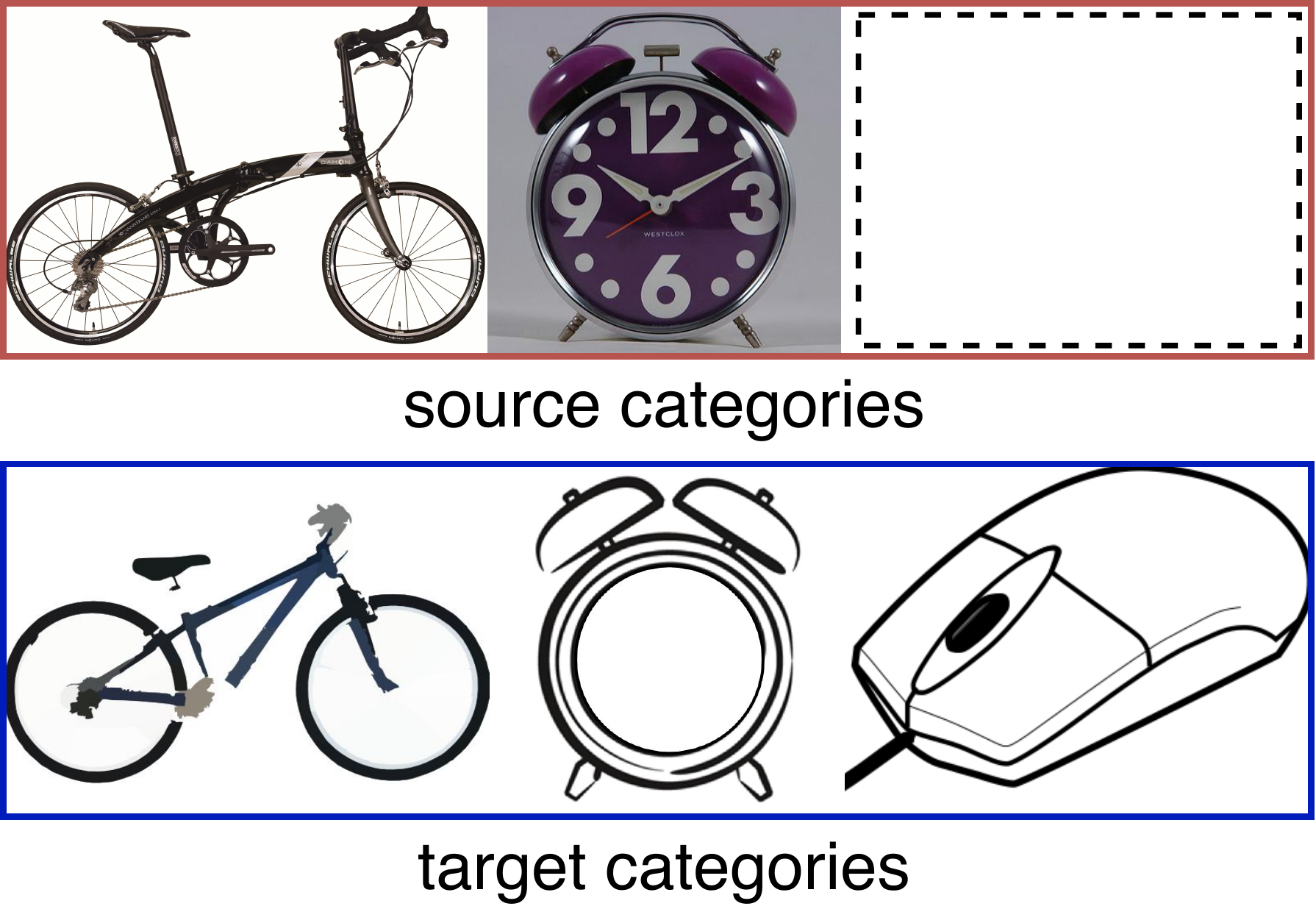}
}
\hfill
\subcaptionbox{Universal Domain Adaptation}{
\includegraphics[width=0.23 \textwidth]{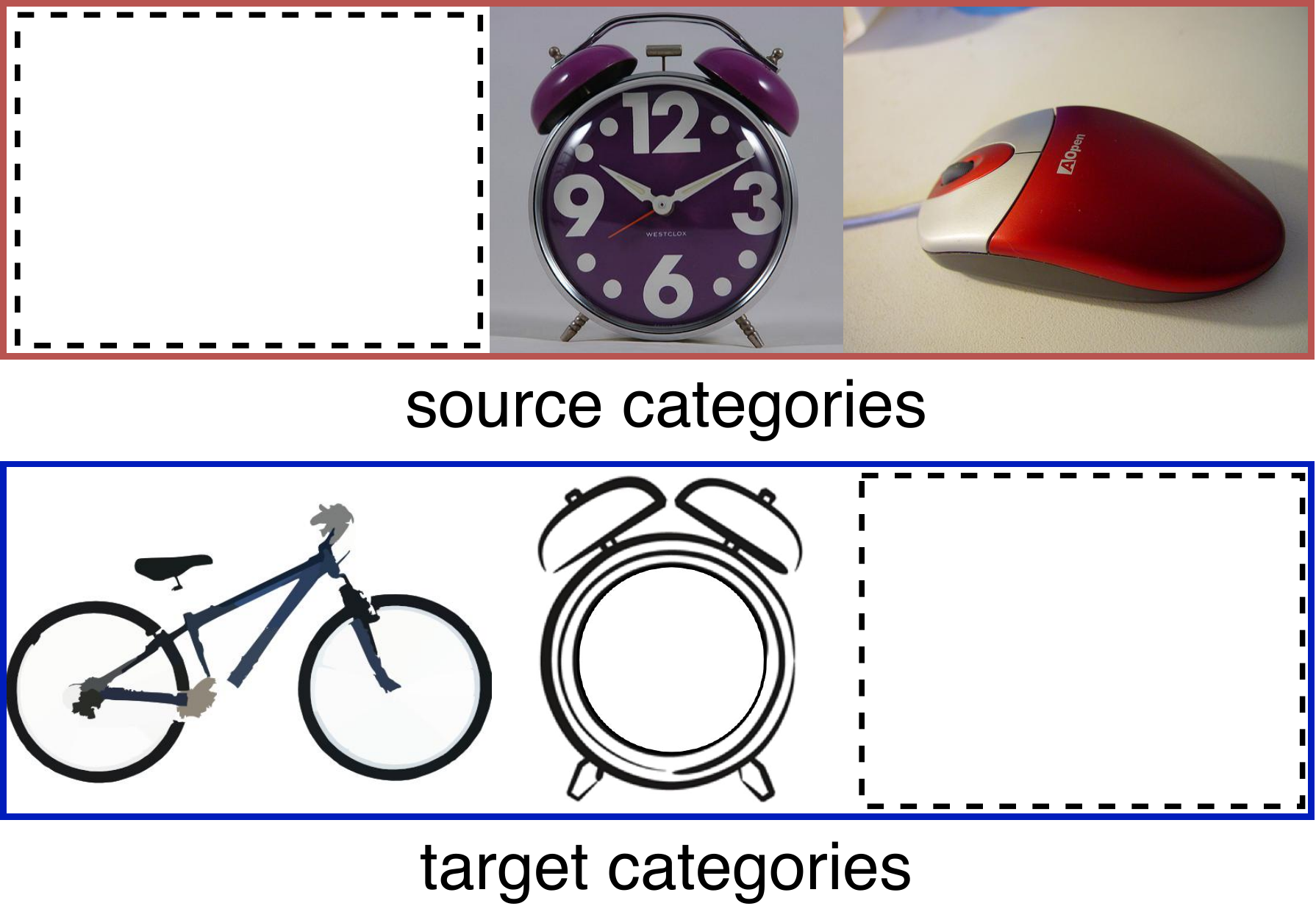}
}
\caption{Four UDA settings with respect to labels of source and target domains. Real pictures in the first row of each figure represent several categories in the source domain and clipart pictures in the second row denote categories in the target domain. A dashed rectangle indicates the absence of categories within one domain, which conversely exists within the other domain.}
 \label{fig:foursettings}
 
 \vskip -0.12in
\end{figure*}

Deep neural networks (DNNs) have demonstrated remarkable advancements across various applications, such as image classification \cite{he2016deep}, object detection \cite{RenHGS15}, and semantic segmentation \cite{LongSD15}. Nevertheless, due to the data-driven nature of DNNs, the pronounced reliance on annotated in-domain data imposes significant constraints on their effectiveness in cross-domain scenarios. To surmount this challenge, Unsupervised Domain Adaptation (UDA) \cite{yang2020transfer} has emerged as a viable remedy by facilitating the transfer of knowledge from a labeled source domain to an unlabeled target domain. UDA has witnessed substantial strides across multiple applications \cite{YuanYB19, DengZES14, ZouYKW18, zheng2023both,zhu2023patch} and most UDA models operate under the Closed Set Domain Adaptation (CSDA) setting, where both domains share an identical label space as shown in Figure \ref{fig:csda_illustration}. However, this assumption often proves impractical in real-world applications. To alleviate the limitations of the CSDA setting, several alternative settings have been proposed, including Partial Domain Adaptation (PDA) \cite{CaoMLW18, guo2022selective}, Open Set Domain Adaptation (OSDA) \cite{SaitoYUH18}, and Universal Domain Adaptation (UniDA) \cite{YouLCWJ19}. Specifically, PDA assumes that the label space of the target domain is a subset of that in the source domain, while OSDA takes the converse assumption that the label space of the source domain is a subset of that in the target domain. UniDA, residing between these extremes, assumes that the label spaces of the two domains exhibit some degree of overlap. As illustrated in Figure \ref{fig:foursettings}, these three settings collectively contribute valuable strides towards enhancing the practical applicability of UDA.

As introduced in Section \ref{sec:related_work}, existing works typically focus on solving each individual UDA setting by proposing distinct methodologies. 
\textit{While those methods have attained remarkable achievements within their specific UDA settings, their effectiveness becomes insufficient within intricate application contexts, necessitating the formulation of a unified framework to address different UDA settings}.

Hence, this study endeavors to construct a unified framework capable of resolving diverse UDA settings. However, this endeavor is met with two pivotal challenges. The first challenge is how to effectively differentiate common classes shared across domains from domain-specific private classes during the cross-domain knowledge transfer and the second one is how to achieve the alignment between two domains characterized by dissimilar label spaces.

To address those challenges, we propose a novel yet effective method called \textbf{L}earning \textbf{I}nstance \textbf{W}eighting for \textbf{U}nsupervised \textbf{D}omain \textbf{A}daptation (\textbf{LIWUDA}) that undertakes the identification and alignment of common classes within two domains. Specifically, the proposed LIWUDA method commences with the design of a weight network to assign weights to individual instances, thereby measuring the probability of instance belonging to common classes. Subsequently, by leveraging these instance weights, we propose the Weighted Optimal Transport (\textbf{WOT}), which effectively minimizes the transportation cost from the source domain to the target domain, facilitating the domain alignment. To separate instances with low similarities and align samples with high similarities, we put forth the Separate and Align (\textbf{SA}) loss, underpinned by a comparison of similarities against the mean cost of WOT. Finally, to guide the learning of the weight network, we introduce the Intra-domain Optimal Transport (\textbf{IOT}) based on WOT to transport one domain, which has only common classes, to itself with a uniform data distribution. In this way, we provide auxiliary information to help learn the weight network. By combining these strategies, the proposed LIWUDA method culminates in a unified framework that comprehensively addresses the four distinctive UDA settings, effectively discovering shared common classes and mitigating domain divergence.

In summary, our main contributions are four-fold. (\textbf{I}) We introduce the unified LIWUDA framework, proficiently addressing four distinct UDA settings. (\textbf{II}) We propose a weight network to assign instance weights, quantifying the probability of an instance belonging to common classes, and subsequently employ this weighting mechanism to devise the WOT strategy, effectively mitigating the domain divergence between the source and target domains. (\textbf{III}) We innovate the SA loss, effectively segregating and aligning samples to harness the cost of WOT, complemented by the IOT, which furnishes auxiliary information for the training of the weight network. (\textbf{IV}) Extensive experiments are conducted to verify the effectiveness of the proposed LIWUDA method across varying UDA settings.

\section{Related Work}
\label{sec:related_work}

\noindent{\bf Closed Set Domain Adaptation}. In CSDA, where the source and target domains are assumed to share an identical label space, the primary research focus lies in mitigating distributional disparities across domains by learning domain-invariant feature representations. This pursuit involves two primary strategies. Specifically, the first strategy capitalizes on statistic moment matching, exemplified by methods such as Maximum Mean Discrepancy (MMD) \cite{LongC0J15, LongZ0J17}, central moment discrepancy \cite{ZellingerGLNS17}, and second-order statistics matching \cite{SunS16}. The second strategy involves adversarial learning paradigms that foster indistinguishability of samples across domains concerning domain labels, typified by domain adversarial neural network and its extensions \cite{GaninUAGLLML16, HoffmanTPZISED18}. \textit{The applicability of CSDA methods to other UDA settings featuring disparate label spaces across domains is not straightforward. Differently, we propose the unified LIWUDA framework, which adeptly accommodates four distinct UDA settings by leveraging the proposed weight network.}

\noindent{\bf Partial Domain Adaptation}. Within the PDA paradigm, the assumption is that the source domain encompasses private classes that does not exist in the target domain, and several methods have been proposed. For instance, PADA \cite{cao2018partial} assigns decreased weight to private classes in the source domain, while IWAN \cite{zhang2018importance} devises a weight assignment mechanism facilitated by an adversarial network to discern samples from private classes. Similarly, SAN \cite{cao2018partial2} employs adversarial learning to identify private classes in the source domain, and ETN \cite{cao2019learning} designs a progressive weighting scheme to quantify the transferability of source instances. DRCN \cite{li2020deep}, on the other hand, introduces a weighted class-wise matching strategy, explicitly aligning target instances with the most relevant source classes. \textit{Diverging from those methodologies, we propose WOT for the purpose of source instance weighting, thereby alleviating the deleterious effects brought by private classes in the source domain.}

% PDA assumes that the source domain contains private classes which are unknown to the target domain and several models are proposed to learn under the PDA setting. For example, Partial Adversarial Domain Adaptation (PADA) \cite{cao2018partial} proposes to down-weigh private classes in the source domain and Importance Weighted Adversarial Nets (IWAN) \cite{zhang2018importance} designs a weighting scheme to detect samples from private classes based on an adversarial network. 
% Similarly, Selective Adversarial Network (SAN) \cite{cao2018partial2} determines source-private classes based on adversarial learning and Example Transfer Network (ETN) \cite{cao2019learning} learns a progressive weighting scheme to quantify the transferability of source instances. Deep Residual Correction Network (DRCN) \cite{li2020deep} proposes a weighted class-wise matching strategy to explicitly align target instances to the most relevant source subclasses. Different from those methods, we propose the WOT to weigh source instances.

\noindent{\bf Open Set Domain Adaptation}. In this work, we mainly focus on the OSDA setting proposed in \cite{SaitoYUH18}, where the target domain encompasses private classes unfamiliar to the source domain. The prevailing objective among OSDA methodologies centers on the identification and exclusion of outliers originating from private classes in the target domain, subsequently reducing the domain gap between source instances and target inliers belonging to common classes. For instance, OSBP \cite{SaitoYUH18} proposes an adversarial learning framework to enable the feature generator to acquire representations conducive to the delineation between common and private classes in the target domain. Likewise, ROS \cite{BucciLT20} employs a self-supervised learning approach, adeptly achieving dual objectives of separating common/private classes in the target domain and aligning domains. \textit{In contrast, the proposed LIWUDA method leverages the weights obtained from the weight network to discriminate whether each target instance is affiliated with private classes.}

% In this work, we mainly follow the OSDA setting proposed by \cite{SaitoYUH18}, where the target domain holds private classes that are unknown to the source domain. Most OSDA models aim to reject outliers from target-private classes and then correctly decrease the domain gap between source instances and target inliers from common classes.  For example, OSBP \cite{SaitoYUH18} proposes an adversarial learning framework that enables the feature generator to learn representations to achieve the separation between common and private classes. ROS  \cite{BucciLT20} employs a self-supervised learning technique to achieve both the common$/$private class separation and the domain alignment. Differently, the proposed LIWUDA model discriminates whether target instances are from private classes based on their weights learned from the weight network.

\noindent{\bf Universal Domain Adaptation}. As a much challenging setting, UniDA introduces a heightened level of complexity by accommodating private classes within both domains. \cite{YouLCWJ19} proposes the UniDA framework, which eschews the necessity for prior knowledge of label space in two domains and introduces the Universal Adaptation Network (UAN) to classify whether each target instance is in common classes. CMU \cite{FuCLW20} combines the entropy, confidence, and consistency metrics to effectively gauge the proclivity of a target instance towards private classes in the target domain, thereby enabling the detection of private classes and accurate classification of data within common classes of the two domains. Similarly, \cite{SaitoKSS20} introduces neighborhood clustering and entropy separation to learn discriminative feature representations for target instances. \textit{Notwithstanding these advancements, all these models overlook the predilection of source instances towards private classes in the source domain, potentially leading to suboptimal performance. In contrast, the proposed LIWUDA model introduces the SA loss, designed to effectively separate instances with low similarities and align instances with high similarities across domains.}

\noindent{\bf Optimal Transport}. Optimal transport (OT) is initially introduced by~\cite{kantorovich2006translocation}, originated as an efficient technique for redistributing mass distributions. In recent years, OT has been harnessed extensively within the domain adaptation domain~\cite{CourtyFTR17, Courty2017JointDO, Redko2016TheoreticalAO, Redko2018OptimalTF, Yan2018SemiSupervisedOT, xu2022few, redko2017theoretical, DamodaranKFTC18, XuLWC020, shen2018wasserstein, Fatras2021UnbalancedMO}, aligning the source and target domains by optimizing the coupling matrix to minimize the cost of transporting from the source domain to the target domain. Among those works, a representative one is DeepJDOT \cite{DamodaranKFTC18}, which employs the learned optimal transport coupling matrix to align two domains in the learned feature space.  Additionally, advancements have led to new optimal transport algorithms such as \cite{courty2014domain}, which incorporates label information through a combination of matrix scaling and non-convex regularization. The SWD method \cite{lee2019sliced} captures the inherent dissimilarities between outputs of task-specific classifiers, enabling the effective detection of target samples distant from the support of the source domain and facilitating end-to-end distributional alignment. RWOT \cite{XuLWC020} designs an integration of shrinking subspace reliability and a discriminative centroid loss. \textit{Different from the aforementioned models that only handle the CSDA setting, we introduce WOT built upon optimal transport, thereby enabling the proposed method to effectively address four distinct scenarios in UDA. Through the utilization of the WOT, we propose the SA loss to successfully separate and align instances, which is bolstered by the IOT loss that imparts auxiliary insights for obtaining weights of instances. As a culmination, the proposed model adeptly tackles the challenges of discriminating common and domain-specific classes, while also accurately quantifying domain divergence.}

% In recent years, Optimal Transport (OT) \cite{CourtyFTR17, DamodaranKFTC18, XuLWC020} has been applied to solve domain adaptation problems, and it aligns the source and target domains by obtaining the optimal coupling and minimizing the cost transported from the source domain to the target domain. 

% Optimal transport has been applied to the CSDA setting to align the source and target domains. For example, Deep Joint Distribution Optimal Transport (DeepJDOT) \cite{DamodaranKFTC18} applies the coupling matrix learned in optimal transport to align two domains in the learned feature space. Reliable Weighted Optimal Transport (RWOT) is designed in \cite{XuLWC020} to learn under the CSDA setting. Joint Class Proportion and Optimal Transport (JCPOT) \cite{RedkoCFT19} performs multi-source adaptation and corrects the distributional shift simultaneously by learning class probabilities of unlabeled target instances and the coupling to align two domains. However, the aforementioned models only handle the CSDA setting.

\begin{figure*}[!htbp]
\begin{center}
%\framebox[4.0in]{$\;$}
\includegraphics[width=0.9\linewidth]{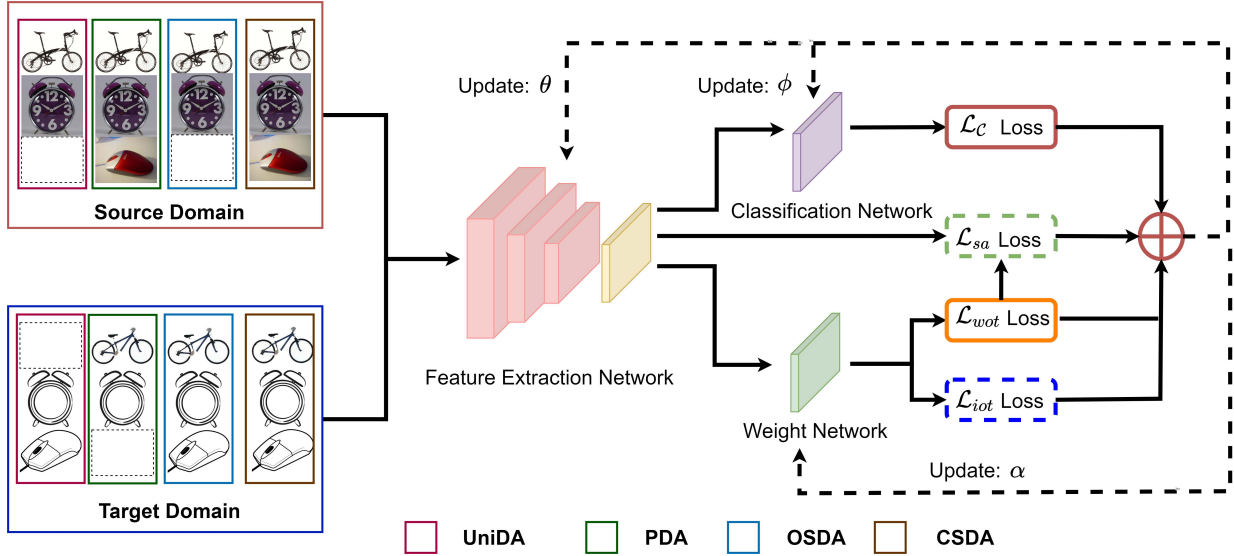}
\caption{Illustration of LIWUDA. Input data varies based on the UDA setting, indicated by different colors representing UniDA, PDA, OSDA, and CSDA. Instance feature representations are obtained through the feature extraction network $f(\cdot;\theta)$. Those representations are then utilized as inputs for both the classification network $h(\cdot;\phi)$ and the weight network $w(\cdot;\alpha)$ to obtain predictions and instance weights. The classification loss $\mathcal{L}_{\mathcal{C}}$ is computed using labeled source instances. By utilizing the feature representations and instance weights, we calculate the cost of WOT $\mathcal{L}_{wot}$. Additionally, we formulate the SA loss $\mathcal{L}_{sa}$ by leveraging the feature representation and $\mathcal{L}_{wot}$. The incorporation of the IOT loss $\mathcal{L}_{iot}$ aids in guiding the training of the weight network. A dashed box denotes potential inapplicability of the corresponding loss in certain settings.} 
\label{fig:framework}
\end{center}
\vskip -0.3in
\end{figure*}

\section{The LIWUDA Method}

In this section, we present the proposed LIWUDA method.

\subsection{Problem Settings}

In UDA, we are given a labeled source domain $D_{s} = \{{(x_{i}^{s},y_{i}^{s})}\}_{i=1}^{n_{s}}$ with $n_{s}$ instances and an unlabeled target domain $D_{t} = \{{{x}_{j}^{t}}\}_{j=1}^{n_{t}}$ with $n_{t}$ unlabeled instances. The label spaces of the source and target domains are respectively denoted by $C_{s}$ and $C_{t}$, with differing data distributions characterizing the source and target domains.

In the UniDA setting, $C = C_{s} \cap C_{t}$ represents the nonempty set of common classes shared by both domains. Meanwhile, $\Bar{C}_{s}= C_{s} \setminus C$ and $\Bar{C}_{t} = C_{t} \setminus C$ denote the sets of private classes exclusive to the source and target domains. The aim of UniDA is to accurately predict target instances in $C$ and adeptly discern target instances within $\Bar{C}_{t}$.

In the context of PDA, the label space $C_{s}$ in the source domain is a superset of $C_{t}$ in the target domain, i.e., $C_{t} \subset C_{s}$. By detecting the set of source-private classes $\Bar{C}_{s} = C_{s} \setminus C_{t}$, the essence of PDA lies in precise prediction for the target domain.

Under the OSDA setting, the label space in the source domain is a subset of that in the target domain, i.e., $C_{s} \subset C_{t}$. Consequently, the target domain introduces a set of private classes $\Bar{C}_{t} = C_{t} \setminus C_{s}$, often referred to as $unknown$ classes, owing to the dearth of prior knowledge. Much like UniDA, OSDA strives for accurate predictions within $C_s$ for target instances, while distinguishing those within $\Bar{C}_{t}$.

In the CSDA setting, both domains share identical label sets $C_{s} = C_{t}$. The core objective revolves around domain alignment, enabling a model trained within the labeled source domain to be effectively employed for the target domain.

\subsection{The Framework}
% \begin{figure*}[!htbp]
% \begin{center}
% %\framebox[4.0in]{$\;$}
% \includegraphics[width=0.99\linewidth]{framework3.pdf}
% \caption{Illustration of the proposed LIWUDA method. The input data depend on the setting we consider, where four colors indicate the four settings, i.e., UniDA, PDA, OSDA, and CSDA. The feature representation of each instance is extracted by the feature extraction network $f(\cdot;\theta)$ and the classification loss $\mathcal{L}_{\mathcal{C}}$ is computed on labeled source instances. After obtaining two ideal couplings $\gamma^*$ and $\kappa^*$, we compute the losses $\mathcal{L}_{wot}$, $\mathcal{L}_{sa}$, and $\mathcal{L}_{iot}$. By minimizing those losses, we can simultaneously align the two domains and weigh instances to distinguish whether each instance comes from common classes or private classes. A dashed box indicates that the corresponding loss may be not used under some settings.}
% \label{fig:framework}
% \end{center}
% % \vskip -0.3in
% \end{figure*}

As illustrated in Figure \ref{fig:framework}, the architecture of the \mbox{LIWUDA} method comprises a triad of pivotal components: a feature extraction network $f(\cdot;\theta)$, a classification network $h(\cdot;\phi)$, and a weight network $w(\cdot;\alpha)$.
The feature extraction network $f(\cdot;\theta)$ parameterized by $\theta$ transforms each data instance into a latent representation. Concurrently, the classification network $h(\cdot;\phi)$ parameterized by $\phi$ leverages the output of $f(\cdot;\theta)$ to form predictions. The weight network $w(\cdot;\alpha)$ characterized by $\alpha$ 
processes the output of $f(\cdot;\theta)$ to generate a scalar weight representing the probability of an instance belonging to the shared common classes in either the source or target domain or both. To confine the output to the range of $[0,1]$, a sigmoid function is applied at the top of the weight network $w(\cdot;\alpha)$. Furthermore, the computation of a normalized output, $\hat{w}(\cdot;\alpha)$, is essential, ensuring that $\sum_{x\in D}\hat{w}(x;\alpha)=1$ within dataset $D$, designated as $D_s$ or $D_t$. For notation simplicity, we omit the dependency of the four functions with respect to model parameters and denote them by $f(\cdot)$, $h(\cdot)$, $w(\cdot)$, and $\hat{w}(\cdot)$.

Based on the architecture introduced above, the LIWUDA method formulates its central objective, characterized by the integration of four distinct loss functions, as
\begin{equation}
\label{eq:lossall}
\mathcal{L} = \mathcal{L}_{\mathcal{C}} + \beta \mathcal{L}_{wot}+ \eta \mathcal{L}_{sa} + \epsilon \mathcal{L}_{iot},
\end{equation}
%\noindent
where $\beta$, $\eta$, and $\epsilon$ are nonnegative hyperparameters. 
%The LIWUDA method aims to minimize $\mathcal{L}$ defined in Eq.~(\ref{eq:lossall}) to learn model parameters $\{\theta,\phi,\alpha\}$.
In the following, we introduce the four losses  %embark on a comprehensive explication of the four losses, meticulously
used in Eq.~\eqref{eq:lossall} in a step-by-step manner.

The first loss $\mathcal{L}_{\mathcal{C}}$ represents the classification loss on the labeled source domain and it is formulated as
\begin{align}
\mathcal{L}_{\mathcal{C}} = \frac{1}{n_{s}}\sum_{i=1}^{{n}_{s}}\ell\left(h(f(x^s_i)),{y}_{i}^{s}\right),\label{classification_loss}
\end{align}
where $h(f(x^s_i))$ denotes the classification probability for the $i$-th source instance $x^s_i$ based on the feature extraction and classification networks, and $\ell (\cdot,\cdot)$ denotes the cross-entropy loss function.

\subsubsection{Weighted Optimal Transport}

In this work, our objective is to introduce a comprehensive framework capable of addressing all four UDA scenarios. However, given the presence of domain-specific classes, direct domain alignment between domains can lead to negative transfer. To mitigate this issue, we introduce a weight network that evaluates the probability of instances belonging to common classes. By leveraging instance weights and optimal transport for UDA \cite{CourtyFTR17, Courty2017JointDO, Redko2016TheoreticalAO, Redko2018OptimalTF, Yan2018SemiSupervisedOT}, we devise a novel approach termed Weighted Optimal Transport (\textbf{WOT}) to effectively measure the domain divergence across all UDA settings. Building upon the acquired feature representations and instance weighting, the WOT cost denoted by $\mathcal{L}_{wot}$ is formulated as
\begin{equation}\label{eq:loss2}
\mathcal{L}_{wot} = {\left\langle {{\gamma}^{*}} , \mathcal{D}({X}^{s},{X}^{t}) \right \rangle}_{F},
\end{equation}
%\noindent
where ${\left\langle\cdot , \cdot\right \rangle}_{F}$ is the Frobenius dot product, $\gamma^*$ is the optimal coupling matrix between the source and target domains, ${X}^{s}$ and ${X}^{t}$ refer to the data in the source and target domains, respectively, and $\mathcal{D}({X}^{s},{X}^{t})$ denotes the cost of transport from $X^{s}$ to $X^{t}$. 
Here $\mathcal{D}({X}^{s},{X}^{t})$ is represented by an $n_s\times n_t$ dissimilarity matrix with the $(i,j)$-th element defined as $ \mathcal{D}\left({x}^{s}_i, {x}^{t}_j\right)=1- \frac{\left\langle f({x}^{s}_i), f({x}^{t}_j) \right\rangle}{\lVert f(x^{s}_i) \rVert_2 {\lVert f(x^{t}_j) \rVert_2}}$, where $\|\cdot\|_2$ denotes the $\ell_2$ norm of a vector. 

% is an $n_s\times n_t$ dissimilarity matrix with the $(i,j)$th entry defined as
% $ \mathcal{D}\left({x}^{s}_i, {x}^{t}_j\right)=1- \frac{\left\langle f({x}^{s}_i), f({x}^{t}_j) \right\rangle}{\lVert f(x^{s}_i) \rVert_2 {\lVert f(x^{t}_j) \rVert_2}}$, where $\|\cdot\|_2$ denotes the $\ell_2$ norm of a vector. 

The optimal coupling matrix $\gamma^*$ in Eq.~\eqref{eq:loss2} is computed through the subsequent Kantorovitch problem \cite{kantorovich2006translocation} as
\begin{equation}\label{eq:hy1}
{\gamma}^{*} =  \mathop{\arg\min}_{\gamma \in \mathcal{B}(p(X^s),p(X^t))}{\left\langle \gamma , \mathcal{D}({X}^{s},{X}^{t}) \right \rangle}_{F},
\end{equation}
%\noindent
where $\bm{1}_d$ is a $d$-dimensional vector of all ones, the superscript $^{\top}$ denotes the transpose operation, $p(X^s)$ defines a discrete probability distribution for source instances in $X^s$ by satisfying $p(X^s)\ge 0$ and $p(X^s)^{\top}\mathbf{1}_{n_s}=1$, $p(X^t)$ is defined similarly for $X^t$, and given $p_1\in\mathbb{R}_+^{n_s}$ and $p_2\in\mathbb{R}_+^{n_t}$ satisfying $p_1^{\top}\mathbf{1}_{n_s}=1$ and $p_2^{\top}\mathbf{1}_{n_t}=1$, $\mathcal{B}(p_1,p_2)$ denotes the set of coupling matrices as 
\begin{equation}
\mathcal{B}(p_1,p_2) = \{\gamma \in \mathbb{R}_{+}^{n_s\times n_t} \mid  \gamma \bm{1}_{n_t} = p_1, \gamma^{\top} \bm{1}_{n_s} = p_2 \}.
\end{equation}
%\noindent
%In the original optimal transport,$p(X^s)=\frac{1}{n_s}$ and $p(X^t)=\frac{1}{n_t}$ 

In the original optimal transport, $p(X^s)=\bm{1}_{n_{s}}/{n_{s}}$ and $p(X^t)=\bm{1}_{n_{t}}/{n_{t}}$ remain constant if there is no prior information regarding the data distribution. In contrast, within the proposed WOT, $p(X^s)$ and $p(X^t)$ can either adopt values from $\hat{w}(\cdot)$ or remain constant, contingent upon the specific UDA setting being addressed, as elucidated in the subsequent section. In this light, it can be inferred that the optimal transport concept emerges as a special instance of the proposed WOT methodology.

\subsubsection{Separate and Align}

While the measurement of domain divergence through the cost computation of WOT is feasible, the presence of domain-private classes may introduce less reliable estimation for the domain divergence. Drawing inspiration from JPOT~\cite{xu2020joint}, we harness the WOT cost (i.e., $\mathcal{L}_{wot}$) as a means to effectively manage the separation and alignment of data across domains. Specifically, we employ $\mathcal{L}_{wot}$ to gauge the distance between each coupled instance pair, thereby guiding the decision to separate or align these instances. When the distance between a coupled pair surpasses $\mathcal{L}_{wot}$, it signifies the infeasibility of transferring knowledge between that pair. Computation of the $(i,j)$-th element within the partial coupling matrix for domain alignment can be defined as
% Although we can measure the domain divergence by calculating the cost of WOT, there may exist domain-specific classes causing the less reliable domain divergence. Inspired by JPOT~\cite{xu2020joint}, we leverage the WOT cost (i.e., $\mathcal{L}{wot}$) to facilitate the separation and alignment of data between domains. Specifically, we assess the distance between each pair of coupled instances with $\mathcal{L}{wot}$ to determine whether these instances should be separated or brought into alignment. If the coupling pair distance is bigger than the cost $\mathcal{L}_{wot}$, it indicates that this coupling pair is nontransferable. The computation of the $(i,j)$th element in the partial coupling matrix for aligning instances can be defined as follows:
% After determining the optimal coupling matrix $\gamma^*$, we utilize the cost of the WOT (i.e., $\mathcal{L}_{wot}$) to separate and align data between domains. Specifically, we compare the distance between each coupling pair with $\mathcal{L}_{wot}$ to discriminate whether such pair of samples should be separated or aligned. The $(i,j)$th entry in the partial coupling matrix to align samples is defined as
\begin{align}
\hat{\gamma}^{*}_{i, j} &=\gamma_{i, j}^{*} \times \pi(x_{i}^{s}, x_{j}^{t}, \mathcal{L}_{wot}), \\
\pi(x_{i}^{s}, x_{j}^{t}, \mathcal{L}_{wot})&=1-\frac{1}{2}\left(1+\operatorname{sgn}\left(\mathcal{D}\left(x_{i}^{s}, x_{j}^{t}\right)-\mathcal{L}_{wot}\right)\right),\nonumber
\end{align}
%\noindent
where $\operatorname{sgn}(x)$ equals 1 when $x>0$, and equals $-1$ otherwise. It is easy to see that $\hat{\gamma}^{*}_{i, j}$ equals 0 if $\mathcal{D}\left(x_{i}^{s}, x_{j}^{t}\right)>\mathcal{L}_{wot}$ and otherwise $\gamma^{*}_{i, j}=1$.
Then, the Separate and Align (\textbf{SA}) loss corresponding to the third loss in Eq.~(\ref{eq:lossall}) for separating and aligning data based on the WOT is formulated as
\begin{equation}\label{eq:lossas}
\mathcal{L}_{sa} = \sum_{i,j} \Delta \gamma_{i, j}^{*}(2-\mathcal{D}({x}^{s}_i,{x}^{t}_j)) + \sum_{i, j} \hat{\gamma}_{i, j}^{*}\mathcal{D}({x}^{s}_i,{x}^{t}_j),
\end{equation}
%\noindent
where $\Delta \gamma_{i, j}^{*}=1-\exp(-(\gamma_{i, j}^{*}-\hat{\gamma}_{i, j}^{*}))$. %denotes the residual set of the ideal partial coupling matrix. 
In Eq.~\eqref{eq:lossas}, the first term is to separate data with large dissimilarities from two domains and the second term is to align data from two domains with small dissimilarities. 

\subsubsection{Intra-domain Optimal Transport}

Furthermore, it becomes evident that target instances (or source instances) within the context of the PDA (or OSDA) setting inherently belong to common classes, and their weights could be approximated to be equal without relying on prior information. This particular insight can be harnessed to guide the learning process of the weight network. To achieve this, we employ the Intra-domain Optimal Transport (\textbf{IOT}) technique within the target domain (or source domain). To elaborate, for a given set of $n$ examples denoted as $X=\{x_i\}_{i=1}^n$, we formulate the IOT loss as
% It is easy to see that target instances (or source instances) under the PDA (or OSDA) setting are all in common classes and their weights could be as equal as possible without prior information. This information could guide the learning of the weight network. To achieve this, we apply the WOT to the target domain (or source domain). Specifically, for a set of $n$ examples $X=\{x_i\}_{i=1}^n$, we define the Intra-domain Optimal Transport (IOT) loss as
\begin{align}
\mathcal{L}_{iot} =  {\left\langle {{\kappa}^{*}} , \mathcal{D}(X,X) \right \rangle}_{F},\label{IOT_loss}
\end{align}
where ${\kappa}^{*}$ is obtained by solving the following problem as
\begin{equation}\label{eq:hy2}
{\kappa}^{*} =  \mathop{\arg\min}_{\kappa \in \mathcal{B}(\hat{w}( X),\bm{1}_{n}/n)}{\left\langle \kappa , \mathcal{D}({X},{X}) \right \rangle}_{F}.
\end{equation}
%\noindent
% So under the PDA setting, $X$ equals $X^t$ and $n$ equals $n_t$, while for the OSDA setting, $X$ equals $X^s$ and $n$ equals $n_s$. By solving problem (\ref{eq:hy2}), we expect the normalized output of the weight network, i.e., $\hat{w}(X)$, to approach $\bm{1}_{n}/n$, which provides some guiding information to train the weight network.

In the PDA setting, we have $X = X^t$ and $n = n_t$, while in the OSDA setting, $X$ corresponds to $X^s$ and $n$ equals $n_s$. Through problem (\ref{eq:hy2}), our objective is to enforce the normalized output of the weight network, which is denoted by $\hat{w}(X)$, towards $\bm{1}_{n}/n$. This problem imparts valuable guiding insights for the effective training of the weight network.

\subsubsection{Entire Objective Function}

In summary, the LIWUDA method is to determine the optimal coupling matrices $\gamma^*$ and $\kappa^*$, alongside the minimization of the objective function $\mathcal{L}$. By amalgamating these critical components, the complete objective function of the proposed LIWUDA approach is formulated as
% To summarize, in the LIWUDA method, we need to determine the coupling matrix $\gamma^*$ and $\kappa^*$ and minimize the loss function $\mathcal{L}$. Putting all the considerations together, the entire objective function of the LIWUDA method is formulated as
\begin{align}\label{eq:final}
\min_{\theta, \phi, \alpha} &\ ~ \mathcal{L} =  \mathcal{L}_{\mathcal{C}} + \beta \mathcal{L}_{wot}+ \eta \mathcal{L}_{sa} + \epsilon \mathcal{L}_{iot} \\
\text{s.t}& \ ~ \gamma^* = \mathop{\arg\min}_{\gamma\in\mathcal{B}(p(X^s),p(X^t))}\mathcal{L}_{wot}\nonumber, \\ 
&\ ~\kappa^* = \mathop{\arg\min}_{\kappa\in \mathcal{B}(\hat{w}( {x}),\bm{1}_{n}/{n})} \mathcal{L}_{iot}. \nonumber
\end{align}
%\noindent
Problem \eqref{eq:final} is a bi-level optimization problem \cite{ColsonMS07} and we can use the Stochastic Gradient Descent (SGD) method or its variants to solve it as \cite{caffarelli2010free,chapel2020partial} did. 

%Problem \eqref{eq:final} is a bi-level optimization problem \cite{ColsonMS07} and we can use the Stochastic Gradient Descent (SGD) method to solve it. Here we can directly obtain $\gamma^*$ and $\kappa^*$ by solving problems \eqref{eq:hy1} and \eqref{eq:hy2} as \cite{caffarelli2010free,chapel2020partial} did.

%However, unlike conventional methods that need to update $\gamma$ and $\kappa$ via gradient descent, here we can directly obtain $\gamma^*$ and $\kappa^*$ by solving problems \eqref{eq:hy1} and \eqref{eq:hy2} as \cite{caffarelli2010free,chapel2020partial} did.

\subsection{Applications to Different UDA Settings}

Problem \eqref{eq:final} presents a general formulation for UDA. However, owning to the 
distinct attributes associated with common classes across the four UDA settings, there can be variations in the configuration of the objective function $\mathcal{L}$ and the ensemble of coupling matrices. In the forthcoming sections, we delve into the specifics of each individual setting.

% Due to different characteristics about common classes under the four UDA settings, the objective function $\mathcal{L}$ and the set of coupling matrices could be different. In the following, we present details for each setting.

\subsubsection{Universal Domain Adaptation} 

In the UniDA setting, it is notable that both the source and target domains contain private classes, thereby instigating the weight network to apportion weights to instances from both domains. This prompts the setting of $p(X^s) = \hat{w}( {X}^{s})$ and $p(X^t) = \hat{w}({X}^{t})$ in problem \eqref{eq:hy1}. Notably, given the absence of a prior knowledge concerning common classes shared between the two domains, the utilization of $\mathcal{L}_{iot}$ to steer the weight network's learning process is omitted, thus setting the hypeparameter $\epsilon = 0$.

% Since we have no prior knowledge about common classes in the two domains, $\mathcal{L}_{iot}$ is not used to guide the learning of the weight network and hence we set $\epsilon =0$.
%the final objective of UniDA can be expressed as $\mathcal{L} = \mathcal{L}_{\mathcal{A}} + \beta \mathcal{L}_{wot}+ \eta \mathcal{L}_{sa}.$

\subsubsection{Partial Domain Adaptation} 

In the PDA setting, it is worthy noting that solely the source domain comprises private classes, thereby confining the weight network to weighing instances exclusively within the source domain. Consequently, it becomes apt to assign $p(X^s) = \hat{w}({X}^{s})$ and $p(X^t) = \bm{1}_{n_{t}}/{n_{t}}$ in relation to problem \eqref{eq:hy1}. Noteworthy also is the fact that, due to the encompassing presence of common classes in the target domain, all instances within this domain are endowed with similar weights. Hence $X$ and $n$ in problem \eqref{eq:hy2} are set to $X^t$ and $n_t$, respectively.

% For the PDA setting, only the source domain contains private classes and the weight network only weighs instances in the source domain. Thus, we set $p(X^s) = \hat{w}( {X}^{s})$ and $p(X^t) = \bm{1}_{n_{t}}/{n_{t}}$ in problem \eqref{eq:hy1}. Moreover, because classes in the target domain are all common classes, all the instances in target domain could have similar weights. Hence $X$ and $n$ in problem \eqref{eq:hy2}
%to determine $\gamma*$ in $\mathcal{L}_{wot}$.
%and we have $\gamma \in \mathcal{B}(\hat{w}( {x}^{s} ; \alpha ),\bm{1}_{n_{t}}/{n_{t}})$ in problem \eqref{eq:hy1}.  are set to $X_t$ and $n_t$, respectively. %$\mathcal{L}_{iot} = {\left\langle {{\kappa}^{*}} , \mathcal{D}(x^t,x^t) \right \rangle}_{F}$ for the target domain and the corresponding set of the coupling matrix of IOT among target domain is $\kappa \in \mathcal{B}(\hat{w}( {x^t} ; \alpha ),\bm{1}_{n_t}/{n_t})$.

\subsubsection{Open Set Domain Adaptation} 

In contrast to the PDA setting, the OSDA configuration takes a converse stance, as the source domain exclusively comprises common classes. As such, it becomes apt to establish $p(X^t) = \hat{w}({X}^{t})$ and $p(X^s) = \bm{1}_{n_{s}}/{n_{s}}$ in connection with problem \eqref{eq:hy1}. Furthermore, for the IOT stipulated by problem \eqref{eq:hy2}, the choice of $X$ and $n$ finds its alignment with $X^s$ and $n_s$, respectively.

% Opposite to the PDA setting, in the OSDA setting, classes in the source domain are all the common classes. Hence, we set $p(X^s) = \bm{1}_{n_{s}}/{n_{s}}$ and $p(X^t) = \hat{w}( {X}^{t})$ in problem \eqref{eq:hy1}. In addition, in problem \eqref{eq:hy2} for the IOT, $X$ and $n$ are set to $X^s$ and $n_s$, respectively.
%we utilize $\mathcal{L}_{iot}$ to make sure weights of instances in the source domain are the same and we have $\mathcal{L}_{iot} =  {\left\langle {{\kappa}^{*}} , \mathcal{D}(x^s,x^s) \right \rangle}_{F}$ and $\kappa \in \mathcal{B}(\hat{w}( {x^s} ; \alpha ),\bm{1}_{n_s}/{n_s})$.

\subsubsection{Closed Set Domain Adaptation}

% The CSDA setting poses comparatively lesser complexity than the other three settings due to the absence of private classes in both domains. Consequently, the requirement for the weight network to weigh instances in both domains becomes redundant. To reflect this, we negate the necessity for $\mathcal{L}_{sa}$ and the IOT by assigning $\eta = 0$ and $\epsilon = 0$ respectively. Consequently, in problem \eqref{eq:hy1}, we establish $p(X^s) = \bm{1}_{n_{s}}/{n_{s}}$ and $p(X^t) = \bm{1}_{n_{t}}/{n_{t}}$, thus simplifying the WOT to the basic optimal transport. Moreover, we introduce a novel approach of transporting mass from the source domain to the target domain intrinsically within the same class, thereby formulating local WOT $\mathcal{L}_{wot}^{m}$ for subdomains corresponding to class $m$. In these instances, $C$ symbolizes the number of classes and $m \in{1,2,..,C}$, while the weights attributed to source and target instances assume the form:
The CSDA setting is easier than the other three settings as there is no private class in both domains. Hence, there is no need to use the weight network to weigh instances in both domains, and we do not need $\mathcal{L}_{sa}$ and IOT by setting $\eta =0$ and $\epsilon=0$. As a consequence, we set $p(X^s) = \bm{1}_{n_{s}}/{n_{s}}$ and $p(X^t) = \bm{1}_{n_{t}}/{n_{t}}$ in problem \eqref{eq:hy1}, making WOT reduce to optimal transport, and the entire LIWUDA method becomes OT-based approaches for UDA (\eg, OTDA \cite{CourtyFTR17} and JDOT~\cite{Courty2017JointDO}). Hence, we do not report experimental results on the CSDA setting in Section \ref{sec:exp}. 
%Furthermore, we propose to transport the mass from the source domain to the target domain within the same class. In the subdomain corresponding to class $m$, where $C$ denotes the number of classes and $m \in\{1,2,..,C\}$, the local WOT $\mathcal{L}_{wot}^{m}$ is calculated with $p(X^{sm}) = {\hat{w}}^{m}( {X}^{s})$ and $p(X^{tm}) = {\hat{w}}^{m}( {X}^{t})$. Specifically, the weights for source and target instances are formulated as
% \begin{equation}
% {\hat{w}}^{m}( {X}^{s})=\frac{{y}^{sm}}{\sum_{l=1}^{n_{s}}{y}_{l}^{sm}}, \ 
% {\hat{w}}^{m}( {X}^{t})=\frac{\hat{y}^{tm}}{\sum_{l=1}^{n_{t}}\hat{y}_{l}^{tm}},  
% \end{equation}
% where ${y}^{s}$ is the one-hot encoding of true labels for the source domain, $\hat{y}^{t}$ is the output of the classification network for target instances, and ${y}^{sm}$ (or ${y}^{tm}$) denotes the $m$-th entry of $y^{s}$ (or  $\hat{y}^{t}$). Finally, $\mathcal{L}_{wot}$ is the sum of $\mathcal{L}_{wot}^{m}$.
%In this case, the proposed LIWUDA method under the CSDA setting is superior to existing optimal transport models \cite{DamodaranKFTC18,CourtyFTR17} based on the results of experiments.
%Moreover, it is not necessary to calculate $\mathcal{L}_{sa}$ to separate and align samples and the proposed LIWUDA for CSDA is similar to existing optimal transport models \cite{DamodaranKFTC18,XuLWC020} for UDA. Therefore, we set $\eta =0$ and $\epsilon=0$ and the final objective for the CSDA setting is $\mathcal{L} = \mathcal{L}_{\mathcal{A}} + \beta \mathcal{L}_{wot}$.

In summary, Table \ref{tab:foursetting} serves to concisely delineate the divergent formulations applied across the four UDA settings. Remarkably, LIWUDA emerges as an overarching framework capable of accommodating all the four UDA settings. The training regimen  of the LIWUDA method is elucidated through Algorithms \ref{Algo: training}. Moreover, for the OSDA and UniDA settings, Algorithm \ref{Algo:Inference} comprehensively delineates the inference process, where the output of the weight network is used to identify whether a sample is from a common or private class.%, thus exemplifying the broader applicability of the approach.

% In Table \ref{tab:foursetting}, we summarize the differences in the formulations for the four UDA settings. The LIWUDA method provides a unified framework for the four settings in UDA. Except the unified formulation for the four settings, another benefit of the unified framework is that we can easily analyze some properties (e.g., generalization bound) of the LIWUDA method under the four settings. In summary, the training procedure and inference process of the LIWUDA method are shown in Algorithms \ref{Algo: training} and \ref{Algo:Inference}, respectively, where the inference process under the OSDA setting is illustrated as an example.

\begin{table}[t]
\centering
\caption{Comparison of formulations for four UDA settings.}
\resizebox{\linewidth}{!}{
\begin{tabular}{lcccccccccccc}
\toprule
\multirow{2}{*}{Setting}&\multirow{2}{*}{$\mathcal{L}_{\mathcal{C}}$}&\multirow{2}{*}{$\mathcal{L}_{wot}$}&\multirow{2}{*}{$\mathcal{L}_{sa}$} &\multirow{2}{*}{$\mathcal{L}_{iot}$}& \multicolumn{2}{c}{Problem (\ref{eq:hy1})}&\multicolumn{2}{c}{Problem (\ref{eq:hy2})} \\
\cmidrule(lr){6-7} \cmidrule(lr){8-9}
&&&&& $p(X^s)$ & $p(X^t)$ & $X$ & $n$\\
\midrule
UniDA&\makecell[c]{$\surd$}&\makecell[c]{$\surd$}&\makecell[c]{$\surd$}&& $\hat{w}({X}^{s})$ & $\hat{w}({X}^{t})$ & - & -\\
PDA&\makecell[c]{$\surd$}&\makecell[c]{$\surd$}&\makecell[c]{$\surd$}&\makecell[c]{$\surd$}&$\hat{w}({X}^{s})$&$\bm{1}_{n_{t}}/{n_{t}}$&$X_{t}$&$n_{t}$\\
OSDA&\makecell[c]{$\surd$}&\makecell[c]{$\surd$}&\makecell[c]{$\surd$}&\makecell[c]{$\surd$}&$\bm{1}_{n_{s}}/{n_{s}}$&$\hat{w}({X}^{t})$&$X_{s}$&$n_{s}$\\
CSDA&\makecell[c]{$\surd$}&\makecell[c]{$\surd$}&&&$\bm{1}_{n_{s}}/{n_{s}}$&$\bm{1}_{n_{t}}/{n_{t}}$& - & -\\
\bottomrule
\end{tabular}}
\label{tab:foursetting}
\end{table}

%\subsection{Algorithm}

% \textcolor{red}{The proposed iterative training procedure is summarized in Algorithm \ref{Algo: training}. In each iteration, the input source and target samples are first input into the feature extractor to output source and target features. After the samples of two domains are mapped to the latent space, we utilize the WOT to obtain the ideal coupling and measure the discrepancy between the two domains. Then we calculate the SA loss based on WOT to separate the features with less similarity and align features with larger similarity. Moreover, we use the prior information that samples with common classes have the same weights. So that the latent distribution is facilitated to be domain-invariant with common classes. In all experiments, we set default values for hyperparameters $\beta$, $\eta$, and $\epsilon$, as 0.1, 0.3, and 0.05 separately. We explore the sensitivity of our proposed method in the sensitivity analysis. The proposed iterative inference procedure is summarized in Algorithm \ref{Algo:Inference}, and we take the OSDA setting as an example. We first obtain the weights of samples based on the weight network and feature extractor. After utilizing the weights to discriminate the samples from common classes or private classes, we make final predictions of samples by the classification network.}

\begin{algorithm}[t]
	\caption{Training process of the LIWUDA framework} 
	\label{Algo: training} 
	\begin{algorithmic}[1]
	    \STATE \textbf{Input}: $\{X,Y\}$; max iterations: $T$
	    %\\ model: $f(\cdot; \theta)$, $h(\cdot; \phi)$, $w(\cdot;\alpha)$;
	    \STATE  Initialize parameters of feature extractor $\theta$, those of the classifier $\phi$, and those of the weight network $\alpha$;
	    \FOR{$t =$ 1 to $T$}
     \STATE Sample a mini-batch;
    	    \STATE Compute the classification loss according to Eq. \eqref{classification_loss};
    	    
    	    \STATE Obtain the idea coupling matrix according to Eq. \eqref{eq:hy1};
    	   
            \STATE Compute the WOT loss according to Eq. \eqref{eq:loss2};
            
            \STATE Compute the SA loss according to Eq. \eqref{eq:lossas};
            
            \STATE Compute the IOT loss according to Eq. \eqref{IOT_loss};
            
            \STATE Compute the overall objective according to Eq. \eqref{eq:lossall};
          
         \STATE Minimize $\mathcal{L}$ to update $\theta$, $\phi$, and $\alpha$;
    	 \ENDFOR
    % 	 \eindent
	    \STATE  \textbf{return}  $\theta$, $\phi$, and $\alpha$.
	\end{algorithmic} 
\end{algorithm}

\begin{algorithm}[t]
	\caption{Inference process of the LIWUDA framework under the \textbf{OSDA} and \textbf{UniDA} settings} 
	\label{Algo:Inference} 
	\begin{algorithmic}[1]
	    \STATE \textbf{Input}: $X$; sample number: $N$
%	    \\ model: $f(\cdot; \theta)$, $h(\cdot; \phi)$, $w(\cdot;\alpha)$;
	    \FOR{$n=1$ to $N$}
	    
    	\STATE Obtain the prediction of a sample: $Y = h(f(x_n))$;
    	    
    	\STATE Obtain the weight of a sample: $W = w(f(x_n))$;
    	    
    	\STATE Predict the sample as common or private class:
     
    	    \quad if  $W>0.5$
    	    
    	    \quad \quad The sample is labeled as common class;
    	       
    	   \quad else
    	   
    	      \quad \quad The sample is labeled as private class;
    	   \STATE Based on $Y$ and common/private class, make final prediction;
    	 \ENDFOR
    % 	 \eindent
	    \STATE  \textbf{return} predictions of samples;
	\end{algorithmic} 
\end{algorithm}

\subsection{Complexity Analysis}
Here, we provide a complexity analysis for the proposed LIWUDA method. Within each iteration, the LIWUDA method necessitates the determination of $\kappa^*$ and $\gamma^*$ by solving the corresponding problems. By using the mini-batch partial optimal transport method~\cite{nguyen2022improving}, solving such a partial optimal transport problem has a computational complexity $\mathcal{O}(n^2)$, where $n$ denotes the number of samples. Then for all the parameters $\Theta = \{\theta,\phi,\alpha\} \in \mathbb{R}^N$, conducting one-step gradient descent to update them costs $\mathcal{O}(N)$. Therefore, the proposed LIWUDA method has a $\mathcal{O}(N+n^2)$ computational cost per iteration, which is as the same order as existing OT-based UDA methods \cite{CourtyFTR17, Fatras2021UnbalancedMO}.

\subsection{Analysis on Generalization Bound}

The LIWUDA method serves as a versatile and unified framework encompassing the four distinct settings within UDA. Beyond its unified formulation for these diverse settings, this framework offers an additional advantage by facilitating the straightforward analysis of critical properties, such as the generalization bound of the LIWUDA method across all four settings. This analysis is the central focus of this section.
% The LIWUDA method provides a unified framework for the four settings in UDA. Except the unified formulation for the four settings, another benefit of the unified framework is that we can easily analyze some properties (e.g., generalization bound) of the LIWUDA method under the four settings, which is the objective of this section.
%Hence, in this section, we provide a theoretical analysis to analyze the generalization bound of the proposed LIWUDA method under the four UDA settings.

The distribution of source domain is represented as $ \mu_s$ on $\Omega$ and the true labeling function in the source domain is denoted by $f_{s}: \Omega \to [0,1]$. Correspondingly, the distribution of the target domain, alongside its inherent true labeling function, can be established in a similar manner. Let's consider a set of functions $h\in\mathcal{H}$ that adhere to the constraint $h: \Omega \to \{0,1\}$, wherein $\mathcal{H}$ denotes a reproducing kernel Hilbert space. Given a convex loss function $l$, we can formally define the expected losses pertinent to both the source and target domains as follows:
\begin{equation}
\begin{aligned}
  R_{S}(h, f_{s}) &= \mathbb{E}_{x\sim \mu_s}[l(h(x),f_s(x))],\\
R_{T}(h, f_{t}) &= \mathbb{E}_{x\sim \mu_t}[l(h(x),f_t(x))].   
\end{aligned}
\end{equation}

In the following theorem, we prove a generalization bound for the proposed LIWUDA method under the four UDA settings. %to upper-bound the target expected loss in terms of the source expected loss based on the Wasserstein distance.

\begin{theorem}
\label{theorem_bound}
Suppose the source domain has $n_{s}$ instances, which are drawn i.i.d from the distribution $\mu_s$ and stored in $X^s$, and the target domain has $n_{t}$ instances, which are drawn i.i.d from the distribution $\mu_t$ and stored in $X^t$. By assuming that the discrete measures of the source and target data are $p(X^s)$ and $p(X^t)$, respectively,  
%for any $d^{\prime}>d$ and $\varsigma^{\prime}<\sqrt{2}$ there exists some constant $N_{0}$ depending on $d^{\prime}$ such that for any $\delta>0$ and $\min \left(n_{s}, n_{t}\right) \geq n_{0} \max \left(\delta^{-\left(d^{\prime}+2\right)}, 1\right)$ with probability at least $1-\delta$ for all $h$ the following holds:
we have 
\begin{equation}
    R_{T}(h,f_{t}) 
\leq  R_{S}(h,f_{s})+W\left(p(X^s), p(X^t)\right)
%\sqrt{\left(\frac{2}{\varsigma^{\prime}}\right) \log \left(\frac{1}{\delta}\right) }\left(\sqrt{\frac{1}{n_{s}}}+\sqrt{\frac{1}{n_{t}}}\right)
+\mathcal{C}+\lambda,
\end{equation}
%\begin{aligned}
%\end{aligned}
where $\mathcal{C}$ is the combined loss of the ideal hypothesis $h^{*}$ that minimizes the combined loss of $R_S(h,f_{s})+ R_T(h,f_{t})$, $\lambda$ denotes the error between the assumed discrete measure and true distribution on both source and target domains, and the Wasserstein distance $W\left(p(X^s), p(X^t)\right)$ is defined as 
\begin{equation*}
W\left(p(X^s), p(X^t)\right) = \min_{\gamma \in \mathcal{B}(p(X^s),p(X^t))}{\left\langle \gamma , \mathcal{D}({X}^{s},{X}^{t}) \right \rangle}_{F}.
\end{equation*}
\end{theorem}

Theorem \ref{theorem_bound} can be directly deduced from Theorem 2 of~\cite{DBLP:conf/pkdd/RedkoHS17} by replacing empirical measures of the source and target domains with $p(X^s)$ and $p(X^t)$, respectively. In the context of traditional optimal transport \cite{DBLP:conf/pkdd/RedkoHS17}, the empirical measures of the source and target domain are conventionally set to $\mathbf{1}_{n_{s}}/{n_{s}}$ and $\mathbf{1}_{n_{t}}/{n_{t}}$, respectively. Nevertheless, within the framework of the LIWUDA method, these measures can be flexibly modified through the updates performed on the weight network. This viewpoint underscores that the LIWUDA method not only endeavors to minimize the divergence between the source and target domains, but also strives to discover improved measures for both domains. To elaborate, let $\bar p(X^{s})$ and $\bar p(X^{t})$ symbolize the optimal discrete measures attributed to the source and target domains, respectively. Consequently, we can establish the following relation:
\begin{equation}
    \begin{aligned}
    W(\bar p(X^{s}), \bar p(X^{t}))& = \min_{\alpha}  W( p(X^{s}),  p(X^{t})) \\
    & \le W(\mathbf{1}_{n_{s}}/{n_{s}},\mathbf{1}_{n_{t}}/{n_{t}}),
\end{aligned}
\end{equation}
where the equality holds due to minimizing $\mathcal{L}_{wot}$ w.r.t $\alpha$ in the LIWUDA method. Note that the above inequality still holds if we fix $p(X^{s})$ as $\mathbf{1}_{n_{s}}/{n_{s}}$ or fix $p(X^{t})$ as $\mathbf{1}_{n_{t}}/{n_{t}}$, and that the right-hand side term of the inequality is just the domain distance in traditional optimal transport. Therefore, the LIWUDA method will have a lower domain distance in all four UDA settings compared with the traditional optimal transport, and this could be a reason of the superior performance of the proposed LIWUDA method under the four UDA settings.

\section{Experiments}
\label{sec:exp}

In this section, we empirically evaluate the proposed \mbox{LIWUDA} method across diverse scenarios including the UniDA, PDA, and OSDA settings.

%Due to that the method of the CSDA setting is a general approach of OTDA, which has been proposed in several previous works, it is not necessary to construct experiments on the CSDA setting. Besides the  settings.

\subsection{Datasets}

We conduct comprehensive experiments on three benchmark datasets: Office-31 \cite{SaenkoKFD10}, Office-Home \cite{VenkateswaraECP17}, and VisDA \cite{abs-1710-06924}.

The Office-31 dataset is a widely-used dataset for UDA, which consists of 4652 images of 31 categories from three domains: DSLR (D), Amazon (A), and Webcam (W). We build six tasks to evaluate our method: {\bf A}$\leftrightarrow${\bf D}, {\bf A}$\leftrightarrow ${\bf W}, and {\bf D}$\leftrightarrow ${\bf W}.
The Office-Home dataset is a more challenging dataset, comprising 15,500 images from 65 classes in 4 distinct domains: Artistic images (Ar), Clip-Art images (CI), Product images (Pr), and Real-World images (Rw). We report the performance of all 12 transfer tasks for comprehensive evaluations: {\bf Ar}$\leftrightarrow${\bf Pr}, {\bf Ar}$\leftrightarrow${\bf Rw}, {\bf Ar}$\leftrightarrow${\bf Rw}, {\bf Cl}$\leftrightarrow${\bf Pr}, {\bf Cl}$\leftrightarrow${\bf Rw}, and {\bf Pr}$\leftrightarrow${\bf Rw}. 
The VisDA dataset, characterized by its large-scale nature, comprises the Synthetic (S) domain containing 152,397 synthetic images and the Real (R) domain with 55,388 images from the real world. By following \cite{LiuCLW019,FuCLW20,LiLLWSHD21}, we construct a task {\bf S}$\rightarrow${\bf R} for the UniDA and OSDA settings, and build two tasks {\bf S}$\leftrightarrow${\bf R} for the PDA setting.

\begin{table}[!bpht]\small
\centering
\caption{The split on the label space, i.e., numbers of common classes (C)/ source-private classes ($\Bar{C}_{s}$)/ target-private classes ($\Bar{C}_{t}$).}
\vskip -0.1in
\begin{tabular}{lccc}
\toprule
\multirow{2}{*}{Dataset}  & \multicolumn{3}{c}{Class split ($\left| C \right|$/$\left|\Bar{C_{s}}\right|$/$\left|\Bar{C_{t}}\right|$)} \\
& OSDA &PDA &UniDA\\
 \midrule
 Office-31  & 10/0/11 & 10/21/0 & 10/10/11\\
 Office-Home & 25/0/40 & 25/40/0  & 10/5/50\\
 VisDA       & 6/0/6 & 6/6/0   & 6/3/3\\
 \bottomrule
 \label{seperation}
 \end{tabular}
\label{tab:split_dat}
\vskip -0.3in
\end{table}

\begin{table*}[t]
\centering
\caption{Results (\%) on the {\bf Office-31} dataset under the {\bf UniDA} setting with ResNet-50 as the backbone.}
\resizebox{\linewidth}{!}{
\begin{tabular}{lccccccccccccccccccccccc}
\toprule
\multirow{2}{*}{Method} & \multicolumn{2}{c}{A$\rightarrow$W} & \multicolumn{2}{c}{D$\rightarrow$W} & \multicolumn{2}{c}{W$\rightarrow$D}& \multicolumn{2}{c}{A$\rightarrow$D} & \multicolumn{2}{c}{D$\rightarrow$A} & \multicolumn{2}{c}{W$\rightarrow$A} & \multicolumn{2}{c}{Avg} \\
 &Acc. & H-score&Acc. & H-score&Acc. & H-score&Acc. & H-score&Acc. & H-score&Acc. & H-score&Acc. & H-score\\
\midrule
PADA \cite{cao2018partial} &85.37& 49.65 &79.26 &52.62 &90.91& 55.60&81.68 &50.00 &55.32 &42.87 &82.61 &49.17 &79.19 &49.98\\
ResNet-50 \cite{he2016deep} & 75.94& 47.92& 89.60 &54.94 &90.91& 55.60&80.45 &49.78& 78.83 &48.48 &81.42 &48.96& 82.86& 50.94\\
ATI \cite{BustoIG20}&79.38 &48.58& 92.60 &55.01& 90.08 &55.45&84.40& 50.48& 78.85& 48.48& 81.57& 48.98& 84.48 &51.16\\
RTN \cite{LongZ0J16} &85.70 &50.21 &87.80& 54.68 &88.91& 55.24&82.69 &50.18 &74.64 &47.65 &83.26 &49.28& 83.83 &51.21\\
IWAN \cite{zhang2018importance}&85.25& 50.13& 90.09& 54.06& 90.00& 55.44&84.27& 50.64& 84.22& 49.65& 86.25& 49.79& 86.68 &51.62\\
OSBP \cite{SaitoYUH18}&66.13 &50.23& 73.57 &55.53 &85.62& 57.20&72.92 &51.14& 47.35 &49.75& 60.48& 50.16 &67.68& 52.34\\
UAN \cite{YouLCWJ19}&85.62 &58.61 &94.77 &70.62& 97.99& 71.42&86.50& 59.68 &85.45 &60.11& 85.12& 60.34& 89.24 &63.46\\
CMU \cite{FuCLW20}&86.86 &67.33& 95.72& 79.32& {\bf 98.01}& 80.42&89.11 &68.11 &88.35& 71.42 &{\bf 88.61} &72.23& 91.11 &73.14\\
% DCC \cite{li2021domain} &91.66 &78.54 &94.52& 79.29 &96.20 &88.58&93.70 &{\bf 88.50} &{\bf 90.43}& 70.18 &{\bf91.97}& 75.87& {\bf 93.08} &80.16\\
\midrule
{\bf LIWUDA} &{\bf 91.84}&{\bf78.91}&{\bf96.28}&{\bf 88.84}&94.74&{\bf 82.14}&{\bf 93.91}&{\bf 82.78}&{\bf 91.67}&{\bf 84.13}&81.53&{\bf 76.28}&{\bf 91.66}&{\bf 82.19}\\
\bottomrule
 \end{tabular}}
\label{tab:office31-universal}
\end{table*}

\begin{table*}[t]
\centering
\caption{H-score (\%) on the {\bf Office-Home} dataset under the {\bf UniDA} setting with ResNet-50 as the backbone.}
\resizebox{\linewidth}{!}{
\begin{tabular}{lcccccccccccccc}
\toprule
 Method & Ar$\rightarrow$Cl & Ar$\rightarrow$Pr & Ar$\rightarrow$Rw & Cl$\rightarrow$Ar & Cl$\rightarrow$Pr & Cl$\rightarrow$Rw & Pr$\rightarrow$Ar & Pr$\rightarrow$Cl & Pr$\rightarrow$Rw & Rw$\rightarrow$Ar & Rw$\rightarrow$Cl & Rw$\rightarrow$Pr & Avg\\
 \midrule
ResNet-50 \cite{he2016deep} & 44.65& 48.04 &50.13 &46.64 &46.91 &48.96& 47.47& 43.17 &50.23 &48.45 &44.76 &48.43& 47.32\\
RTN \cite{LongZ0J16} &38.41& 44.65& 45.70& 42.64& 44.06& 45.48&42.56 &36.79 &45.50 &44.56& 39.79& 44.53& 42.89 \\ 
IWAN \cite{zhang2018importance} & 40.54 &46.96 &47.78 &44.97 &45.06& 47.59&45.81 &41.43 &47.55& 46.29& 42.49& 46.54 &45.25\\
PADA \cite{cao2018partial} & 34.13 &41.89& 44.08& 40.56 &41.52& 43.96&37.04 &32.64 &44.17& 43.06 &35.84& 43.35& 40.19\\
ATI \cite{BustoIG20} &39.88 &45.77& 46.63 &44.13& 44.39& 46.63&44.73 &41.20 &46.59& 45.05 &41.78 &45.45 &44.35\\
OSBP \cite{SaitoYUH18} & 39.59& 45.09& 46.17& 45.70& 45.24 &46.75& 45.26 &40.54 &45.75 &45.08 &41.64 &46.90& 44.48\\
UAN \cite{YouLCWJ19} & 51.64& 51.70& 54.30 &61.74& 57.63& 61.86 &50.38 &47.62& 61.46& 62.87& 52.61& 65.19& 56.58\\
CMU \cite{FuCLW20} &{\bf 56.02}& 56.93& 59.15& {\bf 66.95} &64.27& 67.82&54.72 &{\bf 51.09}&  66.39 & 68.24 &{\bf 57.89}& 69.73 &61.60\\
% DCC \cite{li2021domain} &57.97&54.05&58.01& 74.64 &70.62& 77.52&64.34& 73.60& 74.94& 80.96& 75.12& 80.38& 70.18\\
\midrule
LIWUDA &51.89&{\bf 64.26}&{\bf 66.61}&63.30&{\bf 66.65}&{\bf 69.49}&{\bf 55.00}&47.20&{\bf 71.06}&{\bf 69.67}&55.76&{\bf 72.26}&{\bf 62.76}\\
 \bottomrule
 \end{tabular}
 }
\vskip -0.1in
 \label{tab:officehome-universal}
\end{table*}

\begin{table*}[t]
\vskip -0.1in
\centering
\vspace{0.2cm}
\caption{Accuracy (\%) on the {\bf Office-31} dataset under the {\bf PDA} setting with ResNet-50 as the backbone.}
%\resizebox{\linewidth}{!}{
\begin{tabular}{lccccccc}
\toprule
 Method & A$\rightarrow$W & D$\rightarrow$W & W$\rightarrow$D & A$\rightarrow$D & D$\rightarrow$A & W$\rightarrow$A & Avg \\
 \midrule
DAN \cite{long2015learning} &  59.32$\pm$0.49 & 73.90$\pm$0.38 & 90.45$\pm$0.36 & 61.78$\pm$0.56 & 74.95$\pm$0.67 & 67.64$\pm$0.29 & 71.34 \\
DANN \cite{ganin2016domain} &  73.56$\pm$0.15 & 96.27$\pm$0.26 & 98.73$\pm$0.20 & 81.53$\pm$0.23 & 82.78$\pm$0.18 & 86.12$\pm$0.15 & 86.50 \\
ADDA \cite{tzeng2017adversarial} & 75.67$\pm$0.17 & 95.38$\pm$0.23 & 99.85$\pm$0.12 & 83.41$\pm$0.17 & 83.62$\pm$0.14 & 84.25$\pm$0.13 & 87.03 \\
ResNet-50 \cite{he2016deep} & 75.59$\pm$1.09 & 96.27$\pm$0.85 & 98.09$\pm$0.74 & 83.44$\pm$1.12 & 83.92$\pm$0.95 & 84.97$\pm$0.86 & 87.05 \\
 \midrule
PADA \cite{cao2018partial} &  86.54$\pm$0.31 & 99.32$\pm$0.45 & \textbf{100.0}$\pm$0.00 & 82.17$\pm$0.37 & 92.69$\pm$0.29 & 95.41$\pm$0.33 & 92.69 \\
DRCN \cite{li2020deep} & 88.05 & \textbf{100.0} & \textbf{100.0} & 86.00 & 95.60 & 95.80 & 94.30 \\
IWAN \cite{zhang2018importance} & 89.15$\pm$0.37 & 99.32$\pm$0.32 & 99.36$\pm$0.24 &  90.45$\pm$0.36 & 95.62$\pm$0.29 & 94.26$\pm$0.25 & 94.69 \\
SAN \cite{cao2018partial2} & 93.90$\pm$0.45 & 99.32$\pm$0.52 & 99.36$\pm$0.12 & 94.27$\pm$0.28 & 94.15$\pm$0.36 & 88.73$\pm$0.44 & 94.96 \\
ETN \cite{cao2019learning} & 94.52$\pm$0.20 & \textbf{100.0}$\pm$0.00 & \textbf{100.0}$\pm$0.00 & 95.03$\pm$0.22 & 96.21$\pm$0.27 & 94.64$\pm$0.24 & 96.73 \\
RTNet \cite{chen2020selective} & 96.20$\pm$0.30 & \textbf{100.0}$\pm$0.00 & \textbf{100.0}$\pm$0.00 & 97.60$\pm$0.10 & 92.30$\pm$0.10 & 95.40$\pm$0.10 & 96.90 \\
BA$^3$US \cite{liang2020balanced} & 98.98$\pm$0.28 & \textbf{100.0}$\pm$0.00 & 98.73$\pm$0.00 & 99.36$\pm$0.00 & 94.82$\pm$0.05 & 94.99$\pm$0.08 & 97.81 \\
DCC \cite{li2021domain} & 99.70 & \textbf{100.0} & \textbf{100.0} & 96.10 & 95.30 & 96.30 & 97.90 \\
% SPDA~\cite{guo2022selective}&99.32$\pm$0.02& \textbf{100.0}$\pm$0.00& \textbf{100.0}$\pm$0.00& 96.18$\pm$0.32 &96.03$\pm$0.25 &96.56$\pm$0.00 &98.01\\
\midrule
LIWUDA  & {\bf 99.71}$\pm$0.10 & {\bf 100.00}$\pm$0.00 & {\bf 100.00}$\pm$0.00 & {\bf 99.42}$\pm$0.12 & {\bf 96.33}$\pm$0.15 & {\bf 96.62}$\pm$0.10 & {\bf 98.68}\\
 \bottomrule
 \end{tabular}
%}
% }
 \label{tab:office31-partial}
\end{table*}
\begin{table*}[t]
\centering
\caption{Accuracy (\%) on the {\bf Office-Home} dataset under the {\bf PDA} setting with ResNet-50 as the backbone.}

\resizebox{\linewidth}{!}{
\begin{tabular}{lccccccccccccc}
\toprule
 Method & Ar$\rightarrow$Cl & Ar$\rightarrow$Pr & Ar$\rightarrow$Rw & Cl$\rightarrow$Ar & Cl$\rightarrow$Pr & Cl$\rightarrow$Rw & Pr$\rightarrow$Ar & Pr$\rightarrow$Cl & Pr$\rightarrow$Rw & Rw$\rightarrow$Ar & Rw$\rightarrow$Cl & Rw$\rightarrow$Pr & Avg  \\
 \midrule
DAN \cite{long2015learning} & 35.70 & 52.90 & 63.70 & 45.00 & 51.70 & 49.30 & 42.40 & 31.50 & 68.70 & 59.70 & 34.60 & 67.80 & 50.30  \\
ResNet-50 \cite{he2016deep} & 46.33 & 67.51 & 75.87 & 59.14 & 59.94 & 62.73 & 58.22 & 41.79 & 74.88 & 67.40 & 48.18 & 74.17 & 61.35 \\
DANN \cite{ganin2016domain} & 43.76 & 67.90 & 77.47 & 63.73 & 58.99 & 67.59 & 56.84 & 37.07 & 76.37 & 69.15 & 44.30 & 77.48 & 61.72 \\
ADDA \cite{tzeng2017adversarial} & 45.23 & 68.79 & 79.21 & 64.56 & 60.01 & 68.29 & 57.56 & 38.89 & 77.45 & 70.28 & 45.23 & 78.32 & 62.82 \\
\cmidrule(){1-14}
PADA \cite{cao2018partial} & 51.95 & 67.00 & 78.74 & 52.16 & 53.78 & 59.03 & 52.61 & 43.22 & 78.79 & 73.73 & 56.6 & 77.09 & 62.06 \\
IWAN \cite{zhang2018importance} & 53.94 & 54.45 & 78.12 & 61.31 & 47.95 & 63.32 & 54.17 & 52.02 & 81.28 & 76.46 & 56.75 & 82.90 & 63.56 \\
SAN \cite{cao2018partial2} & 44.42 & 68.68 & 74.60 & 67.49 & 64.99 & 77.80 & 59.78 & 44.72 & 80.07 & 72.18 & 50.21 & 78.66 & 65.30 \\
DRCN \cite{li2020deep} & 54.00 & 76.40 & 83.00 & 62.10 & 64.50 & 71.00 & 70.80 & 49.80 & 80.50 & 77.50 & 59.10 & 79.90 & 69.00 \\
ETN \cite{cao2019learning} & 59.24 & 77.03 & 79.54 & 62.92 & 65.73 & 75.01 & 68.29 & 55.37 & 84.37 & 75.72 & 57.66 & 84.54 & 70.45 \\
RTNet \cite{chen2020selective} & {\bf 63.20}$\pm$0.10 & 80.10$\pm$0.20 & 80.70$\pm$0.10 & 66.70$\pm$0.10 & 69.30$\pm$0.20 & 77.20$\pm$0.20 & 71.60$\pm$0.30 & 53.90$\pm$0.30 & 84.60$\pm$0.10 & 77.40$\pm$0.20 & 57.90$\pm$0.30 & 85.50$\pm$0.10 & 72.30 \\
DCC \cite{li2021domain} & 59.00 & 84.40 & 83.40 & 67.80 & 72.70 & 79.80 & 68.40 & 53.20 & 83.70 & 75.80 & 59.00 & \textbf{88.30} & 73.00 \\
BA$^3$US \cite{liang2020balanced} & 60.62$\pm$0.45 & 83.16$\pm$0.12 & {\bf 88.39}$\pm$0.19 & {\bf 71.75}$\pm$0.19 & 72.79$\pm$0.19 & \textbf{83.40}$\pm$0.59 & 75.45$\pm$0.19 & 61.59$\pm$0.37 & \textbf{86.53}$\pm$0.22 & 79.25$\pm$0.65 & 62.80$\pm$0.51 & 86.05$\pm$0.26 & 75.98 \\

\cmidrule{1-14}
LIWUDA & 62.15$\pm$0.15 & \textbf{84.59}$\pm$0.15 & 85.48 $\pm$0.14 & 71.17$\pm$0.13 & \textbf{74.34}$\pm$0.13 & 80.89$\pm$0.16& {\bf 78.15}$\pm$0.11 & {\bf 64.72}$\pm$0.21 & 84.65$\pm$0.24 & 82.60$\pm$0.12 & \textbf{69.43}$\pm$0.23 & {\bf 87.5}$\pm$0.18 & \textbf{77.14} \\
 \bottomrule
 \end{tabular}
 }
 \label{tab:officehome-partial}
\end{table*}

\subsection{Experimental Setup}
\subsubsection{Constructions of Common/Private Classes}
Following prior works \cite{BustoIG20, SaitoYUH18, YouLCWJ19,CaoMLW18}, we categorize the label into three distinct parts: common classes $C$, source-private classes $\Bar{C}_{s}$, and target-private classes $\Bar{C}_{t}$. The division of three datasets under different settings is depicted in Table \ref{tab:split_dat}. Note that the classes are sorted according to the alphabetical order during the split by following \cite{BustoIG20, SaitoYUH18, YouLCWJ19, CaoMLW18}.

\subsubsection{Evaluations} Under the OSDA and UniDA settings, target-private classes will be merged into a single \emph{unknown} class. Performance is evaluated using two metrics: accuracy ({\bf Acc.}) and {\bf H-score} \cite{BucciLT20}. The former represents the mean per-class accuracy over common classes and the accuracy of the \emph{unknown} (UNK) class. The latter is the harmonic mean of the accuracy for samples with common classes and the \emph{unknown} class, following the prior work \cite{FuCLW20, BucciLT20}.
On the VisDA dataset under the OSDA setting, in line with the previous work \cite{SaitoYUH18, LiuCLW019}, we employ two metrics: {\bf OS} (Overall Accuracy) and {${\mathbf{OS}}^{\mathbf{}}$}, where {\bf OS} is the accuracy, and {${\mathbf{OS}}^{\mathbf{}}$} calculates the mean accuracy exclusively for common classes. Under the PDA setting, we report the average per-class accuracy specifically for common classes.

% , and we measure the performance via two metrics, i.e., the accuracy ({\bf Acc.}) and {\bf H-score} \cite{BucciLT20}, where the former is the mean of per-class accuracy over common classes and the accuracy of the \emph{unknown} (UNK) class, and the latter is the harmonic mean on the accuracy of samples with common classes and \emph{unknown} class by following \cite{FuCLW20,BucciLT20} . On the VisDA dataset under the OSDA setting, by following \cite{SaitoYUH18,LiuCLW019}, we use two metrics: {\bf OS} and {${\mathbf{OS}}^{\mathbf{*}}$}, where {\bf OS} is just the accuracy and {${\mathbf{OS}}^{\mathbf{*}}$} only calculates the mean accuracy on common classes. Under the PDA setting, we report the average per-class accuracy over common classes.

% and so there are namely total 256 instances from the source and target domains.

\subsubsection{Implementation Details}

We adopt ResNet-50 \cite{he2016deep} pretrained on the ImageNet dataset as the feature extraction network. The classification network and weight network are both implemented using single-layer fully connected neural networks. To ensure fair and equitable benchmarking against prior works, the LIWUDA framework for the OSDA setting on the VisDA dataset utilizes the VGG-19 network \cite{SimonyanZ14a} as its foundational backbone. The optimizer is the SGD with a momentum of 0.9, an initial learning rate of 0.001, and a weight decay of 0.005. The batch size for each domain is set to 128. The confidence threshold, which predicts samples as common classes or private classes, is set to 0.5 during the inference process. Specifically, an instance is classified into common classes when the weight of the instance exceeds 0.5; otherwise, it is assigned to private classes. The default values for hyperparameters, including $\beta$, $\eta$, and $\epsilon$ in Eq.~\eqref{eq:lossall}, are set to 0.1, 0.3, and 0.05, respectively, across all settings. For the UniDA setting, $\epsilon$ is set to 0.

% and for the CSDA setting, both $\eta$ and $\epsilon$ are set to 0.

\subsubsection{Baselines} 
%In this paper, we compare our proposed method with previous state-of-the-art works in three settings of domain adaptation, i.e., UniDA, PDA, and OSDA. 
For the UniDA setting, the proposed LIWUDA method is compared with RTN \cite{LongZ0J16}, IWAN \cite{zhang2018importance}, PADA \cite{cao2018partial}, ATI \cite{BustoIG20}, OSBP \cite{SaitoYUH18}, UAN \cite{YouLCWJ19}, CMU \cite{FuCLW20}, and USFDA \cite{KunduVVB20}.
For the PDA setting, LIWUDA is compared with PADA \cite{cao2018partial}, DRCN\cite{LiLLWSHD21}, IWAN \cite{zhang2018importance}, SAN \cite{cao2018partial2}, ETN \cite{cao2019learning}, RTNet \cite{chen2020selective}, BA$^3$US \cite{liang2020balanced}, and DCC~\cite{li2021domain}. In the OSDA setting, comparisons are made with UAN \cite{YouLCWJ19}, STA \cite{LiuCLW019}, OSBP \cite{SaitoYUH18}, ROS \cite{BucciLT20}, and OSVM \cite{JainSB14}, along with two variants MMD+OSVM \cite{TzengHZSD14} and DANN+OSVM \cite{GaninL15}.

%It worth noting that MMD+OSVM and DANN+OSVM are two variants of OSVM that incorporate Maximum Mean Discrepancy \cite{TzengHZSD14} and domain adversarial network \cite{GaninL15}.
%For the CSDA setting, we compare our proposed method with optimal transport methods, including OTDA \cite{DamodaranKFTC18}, JDOT~\cite{Courty2017JointDO}, and DeepJDOT \cite{CourtyFTR17}, and some non-optimal transport based methods such as DGA-DA~\cite{Luo2017DiscriminativeAG} and ARG-DA~\cite{Luo2022AttentionRL}.

\subsection{Experimental Results}

\begin{table*}[!htbp]
%\vskip -0.1in
\centering
\vspace{0.2cm}
\caption{Results (\%) on the {\bf Office-31} dataset under the {\bf OSDA} setting with ResNet-50 as the backbone.}
\resizebox{\linewidth}{!}{
\begin{tabular}{lccccccccccccccccccccccc}
\toprule
\multirow{2}{*}{Method} & \multicolumn{3}{c}{A$\rightarrow$W} & \multicolumn{3}{c}{A$\rightarrow$D} & \multicolumn{3}{c}{D$\rightarrow$W}& \multicolumn{3}{c}{W$\rightarrow$D} & \multicolumn{3}{c}{D$\rightarrow$A} & \multicolumn{3}{c}{W$\rightarrow$A} & \multicolumn{3}{c}{Avg} \\
 &OS$^{*}$ & UNK & H-score &OS$^{*}$ & UNK & H-score&OS$^{*}$ & UNK & H-score&OS$^{*}$ & UNK & H-score &OS$^{*}$ & UNK & H-score&OS$^{*}$ & UNK & H-score&OS$^{*}$ & UNK & H-score\\
\midrule
UAN \cite{YouLCWJ19} &{\bf 95.5}& 31.0 &46.8 &95.6 &24.4 &38.9 &{\bf 99.8} &52.5 &68.8& 81.5 &41.4& 53.0 &{\bf 93.5}& 53.4& 68.0& {\bf 94.1}& 38.8& 54.9 &93.4& 40.3 &55.1\\
STA$_{max}$ \cite{LiuCLW019} & 86.7& 67.6& 75.9& 91.0 &63.9& 75.0& 94.1 &55.5& 69.8&84.9& 67.8& 75.2& 83.1 &65.9& 73.2& 66.2& 68.0& 66.1& 84.3& 64.8 &72.5 \\
OSBP \cite{SaitoYUH18}  & 86.8& 79.2 &82.7& 90.5& 75.5& 82.4 &97.7 &{\bf 96.7}& {\bf 97.2}&99.1& 84.2 &91.1 &76.1& 72.3& 75.1& 73.0 &74.4 &73.7& 87.2 &80.4 &83.7\\
ROS \cite{BucciLT20}  &88.4 &76.7 &82.1& 87.5 &77.8 &82.4 &99.3& 93.0&96.0&{\bf 100.0} &{\bf 99.4}& {\bf 99.7} &74.8& 81.2& {\bf 77.9} &69.7& {\bf 86.6} &77.2& 86.6& {\bf 85.8} &85.9 \\
% DCC \cite{li2021domain}  &-&-&87.1&-&-& 85.5 &-&-&91.2&-&-&87.1 &-&-&{\bf 85.5}&-&-& {\bf 84.4}&-&-&{\bf 86.8}\\
\midrule
LIWUDA &89.8&{\bf86.7}&{\bf88.2}& {\bf99.2}&{\bf83.1}&{\bf90.4}&94.4&91.6&93.0&94.9&86.2&90.3&70.3& {\bf 85.2}&77.0&84.1&71.6& {\bf 77.3} &{\bf 88.8}&84.1&{\bf 86.1}\\
\bottomrule 
 \end{tabular}
 }
%  \vskip 0.1in
\label{tab:office31-openset}
\end{table*}
\begin{table*}[!htbp]
\centering
\caption{Results (\%) on the {\bf Office-Home} dataset under the {\bf OSDA} setting with ResNet-50 as the backbone.}
\resizebox{\linewidth}{!}{
\begin{tabular}{lccccccccccccccccccccc}
\toprule
 \multirow{2}{*}{Method} & \multicolumn{3}{c}{Ar$\rightarrow$Cl} & \multicolumn{3}{c}{Ar$\rightarrow$Pr} & \multicolumn{3}{c}{Ar$\rightarrow$Rw} & \multicolumn{3}{c}{Cl$\rightarrow$Ar}& \multicolumn{3}{c}{Cl$\rightarrow$Pr} & \multicolumn{3}{c}{Cl$\rightarrow$Rw}\\
 &OS$^{*}$ & UNK & H-score &OS$^{*}$ & UNK & H-score&OS$^{*}$ & UNK & H-score &OS$^{*}$ & UNK & H-score&OS$^{*}$ & UNK & H-score&OS$^{*}$ & UNK & H-score\\
 \midrule
UAN \cite{YouLCWJ19}      &{\bf 62.4}& 0.0& 0.0& {\bf 81.1} &0.0& 0.0& {\bf 88.2}& 0.1& 0.2&{\bf  70.5}& 0.0 &0.0 &{\bf74.0} &0.1& 0.2& {\bf80.6}& 0.1 &0.2 \\   
STA$_{max}$ \cite{LiuCLW019}  &46.0& 72.3& 55.8 &68.0 &48.4& 54.0 &78.6 &60.4 &68.3& 51.4& 65.0& 57.4&61.8 &59.1& 60.4 &67.0& 66.7& 66.8 \\ 
OSBP \cite{SaitoYUH18}     &50.2 &61.1& 55.1 &71.8& 59.8 &65.2& 79.3& 67.5 &72.9& 59.4& 70.3& 64.3& 67.0& 62.7& 64.7 &72.0 &69.2& {\bf70.6} \\ 
% DCC \cite{li2021domain} &-&-&52.9&-&-& 67.4&-&-& {\bf 80.6} &-& 49.8&-&-&66.6&-&-& 67.0&-\\
ROS \cite{BucciLT20}     &50.6& {\bf 74.1}& {\bf 60.1}& 68.4& {\bf 70.3}& {\bf 69.3} &75.8& {\bf 77.2}& {\bf 76.5} &53.6& 65.5& 58.9&59.8& {\bf71.6} &65.2&  65.3 &72.2& 68.6\\  
\midrule
LIWUDA &52.4&69.3&59.7&65.2&65.0&65.0&68.2&66.9&67.5&56.2&{\bf 91.9}&{\bf 69.8}&65.3& 70.0&{\bf67.6}&61.1&{\bf73.1}&66.6\\
 \bottomrule
 \bottomrule
 \multirow{2}{*}{Method} & \multicolumn{3}{c}{Pr$\rightarrow$Ar} & \multicolumn{3}{c}{Pr$\rightarrow$Cl}& \multicolumn{3}{c}{Pr$\rightarrow$Rw} & \multicolumn{3}{c}{Rw$\rightarrow$Ar} & \multicolumn{3}{c}{Rw$\rightarrow$Cl} & \multicolumn{3}{c}{Rw$\rightarrow$Pr} & \multicolumn{3}{c}{Avg}\\
 &OS$^{*}$ & UNK & H-score &OS$^{*}$ & UNK & H-score&OS$^{*}$ & UNK & H-score &OS$^{*}$ & UNK & H-score&OS$^{*}$ & UNK & H-score&OS$^{*}$ & UNK & H-score&OS$^{*}$ & UNK & H-score \\
 \midrule
UAN \cite{YouLCWJ19}      &{\bf73.7}& 0.0& 0.0 &{\bf59.1}& 0.0& 0.0&{\bf84.0} &0.1 &0.2 &{\bf77.5}& 0.1& 0.2& {\bf66.2}& 0.0& 0.0& {\bf85.0}& 0.1 &0.1 &{\bf75.2}& 0.0 &0.1 \\   
STA$_{max}$ \cite{LiuCLW019} &54.2 &{\bf72.4}& 61.9& 44.2 &67.1& 53.2 &76.2& 64.3& 69.5& 67.5& 66.7& 67.1 &49.9& 61.1 &54.5& 77.1 &55.4& 64.5 &61.8 &63.3& 61.1 \\ 
OSBP \cite{SaitoYUH18}     & 59.1 &68.1& 63.2& 44.5& 66.3 &53.2&76.2& 71.7& 73.9 &66.1& 67.3& 66.7 &48.0& 63.0 &54.5& 76.3 &68.6& 72.3& 64.1& 66.3& 64.7 \\  
% DCC \cite{li2021domain} &-&-& 59.5&-& 52.8&-&-&64.0&-&-& 56.0&-&-& 76.9&-&-& 62.7&-&-&64.2&- \\ 
ROS \cite{BucciLT20}     & 57.3& 64.3 &60.6& 46.5& {\bf71.2}& 56.3&70.8&{\bf 78.4} &{\bf74.4}& 67.0 &70.8 &{\bf68.8}&   51.5 &{\bf73.0}& 60.4&   72.0& {\bf80.0} &{\bf75.7} &  61.6 &72.4 &66.2       \\  
\midrule
LIWUDA &57.7&71.8&{\bf 63.9}&52.8&70.9&{\bf60.5}&64.6&76.8&70.2&64.5&{\bf 70.9}&67.5&61.5&72.2&{\bf66.4}&70.6&71.4&71.0&61.7&{\bf 72.6}&{\bf 66.3}\\
 \bottomrule
 \end{tabular}
 }
 \label{tab:officehome-openset}
\end{table*}

\begin{table*}[!htbp]
\centering

\caption{Results (\%) on the {\bf VisDA} under three settings: {\bf UniDA} (ResNet-50), {\bf PDA} (ResNet-50), and {\bf OSDA} (VGG-19). %The reported performance for UniDA, PDA, and OSDA is from \cite{FuCLW20}.
}
%\resizebox{\linewidth}{!}{
\begin{tabular}{lccccccccccccccccccc}
\toprule
\multirow{2}{*}{{\bf Method}}&\multicolumn{2}{c}{UniDA} & \multirow{2}{*}{{\bf Method}}&\multicolumn{3}{c}{PDA} & \multirow{2}{*}{{\bf Method}}&\multicolumn{2}{c}{OSDA}\\
 &Acc. & H-score& &R $\rightarrow$ S &S $\rightarrow$ R &Avg& & OS & OS$^{*}$ \\
\midrule
RTN \cite{LongZ0J16} &53.92 &26.02&DAN \cite{long2015learning}&68.35&47.60&57.98& OSVM \cite{JainSB14} & 52.5 &54.9\\
IWAN \cite{zhang2018importance} &58.72 &27.64&DANN \cite{ganin2016domain}&73.84&51.01&62.43&MMD+OSVM \cite{TzengHZSD14} &54.4 &56.0\\
ATI \cite{BustoIG20}&54.81& 26.34&RTN \cite{LongZ0J16} & 72.93&50.04&61.49&DANN+OSVM \cite{GaninL15}& 55.5& 57.8\\
OSBP \cite{SaitoYUH18}&30.26& 27.31&PADA \cite{cao2018partial}&{\bf 76.50}&53.53&65.01& ATI-$\lambda$ \cite{BustoIG20} &59.9 &59.0\\
UAN \cite{YouLCWJ19} &60.83& 30.47&SAN \cite{cao2018partial2}&69.70&49.90&59.80&OSBP \cite{SaitoYUH18}& 62.9 &59.2\\
USFDA \cite{KunduVVB20} & 63.92& -&IWAN \cite{zhang2018importance}&71.30&48.60&60.00&STA \cite{LiuCLW019} &66.8& 63.9\\
CMU \cite{FuCLW20}&61.42& 34.64&DRCN \cite{LiLLWSHD21}&73.20&58.20&65.70&USFDA \cite{KunduVVB20} & {\bf 68.1}&64.7\\
% DCC \cite{li2021domain} &68.8&68.0&SLM \cite{abs-2012-03358}&77.84&91.74&84.61&DCC \cite{li2021domain}&64.20 &43.02\\
\midrule
{\bf LIWUDA}&{\bf 64.25}&{\bf 38.69}&{\bf LIWUDA}&72.93&{\bf 76.73} &{\bf 72.50}&{\bf LIWUDA}&58.6&{\bf 68.4}&\\
\bottomrule
\end{tabular}%}
\label{tab:visda}
\end{table*}

\subsubsection{Results for UniDA} 

In the most challenging UniDA setting, the comparative analysis of results presented in Tables \ref{tab:office31-universal}, \ref{tab:officehome-universal}, and \ref{tab:visda} demonstrates the superior effectiveness of the proposed LIWUDA method over state-of-the-art alternatives across all datasets. On the Office-31 dataset, the proposed LIWUDA showcases a stronger capability to discriminate between unknown classes and known classes. Notably, on the Office-31 dataset, the proposed \mbox{LIWUDA} method attains an outstanding average performance, boasting a substantial \textbf{9\%} enhancement in terms of the H-score. Moreover, it consistently achieves the best performance across the majority of tasks. Similarly, on the Office-Home dataset, the proposed LIWUDA method secures the paramount position in terms of average performance and excels in {\bf 8 out of 12} transfer tasks, while clinching commendable second or third positions in the remaining 4 transfer tasks. A particularly noteworthy achievement is the significant \textbf{1.16\%} increase in H-score compared to the CMU method~\cite{FuCLW20}. It is worth noting that in scenarios where the target domain involves Clipart, the performance of the proposed method experiences a slight dip, attributed partly to the increased complexity and inclusion of noisy data in the Clipart domain. In the VisDA dataset, the proposed \mbox{LIWUDA} methodology emerges as the frontrunner in terms of both evaluation metrics, manifesting a substantial enhancement of \textbf{2.83\%} accuracy and \textbf{4.05\%} in H-score compared to the CMU~\cite{FuCLW20} method.
\begin{table*}[t]
\vskip -0.1in
\centering
\caption{Sensitivity to $\beta$ on the {\bf Office-Home} dataset under the {\bf PDA} setting with ResNet-50 as the backbone.}
\resizebox{\linewidth}{!}{
\begin{tabular}{l|cccccccccccccc}
\toprule
 $\beta$ & Ar$\rightarrow$Cl & Ar$\rightarrow$Pr & Ar$\rightarrow$Rw & Cl$\rightarrow$Ar & Cl$\rightarrow$Pr & Cl$\rightarrow$Rw & Pr$\rightarrow$Ar & Pr$\rightarrow$Cl & Pr$\rightarrow$Rw & Rw$\rightarrow$Ar & Rw$\rightarrow$Cl & Rw$\rightarrow$Pr & Avg\\
\midrule
\midrule
0 &46.33 & 67.51 & 75.87 & 59.14 & 59.94 & 62.73 & 58.22 & 41.79 & 74.88 & 67.40 & 48.18 & 74.17 & 61.35\\
0.05&54.32&78.83&82.12&66.93&66.43&\textbf{78.32}&74.37&53.68&80.81&79.56&63.21&82.46&71.75\\
0.1& \textbf{58.63}&\textbf{79.64}&\textbf{83.10}&67.40&\textbf{68.46}&76.42&\textbf{75.48}&\textbf{57.73}&\textbf{82.33}&\textbf{83.10}&\textbf{65.97}&84.09&\textbf{73.53}  \\
0.2& 57.46&78.43&82.34&\textbf{68.12}&67.82&75.41&73.47&56.42&81.23&80.09&65.43&\textbf{84.21}& 72.54 \\
0.5&57.23&77.34&81.29&66.65&66.73&76.01&75.41& 57.01&80.32&81.27&65.12&83.45&72.32\\
 \bottomrule
 \end{tabular}
 }
\vskip -0.1in
 \label{tab:beta}
\end{table*}
\begin{table*}[t]
\centering
\caption{Sensitivity to $\eta$ on the {\bf Office-Home} dataset under the {\bf PDA} setting with ResNet-50 as the backbone.}
\resizebox{\linewidth}{!}{
\begin{tabular}{l|cccccccccccccc}
\toprule
 $\eta$ & Ar$\rightarrow$Cl & Ar$\rightarrow$Pr & Ar$\rightarrow$Rw & Cl$\rightarrow$Ar & Cl$\rightarrow$Pr & Cl$\rightarrow$Rw & Pr$\rightarrow$Ar & Pr$\rightarrow$Cl & Pr$\rightarrow$Rw & Rw$\rightarrow$Ar & Rw$\rightarrow$Cl & Rw$\rightarrow$Pr & Avg\\
\midrule
\midrule
0 & 58.63&79.64&83.10&67.40&68.46&76.42&75.48&57.73&82.33&\textbf{83.10}&65.97&84.09&73.53 \\
0.1&58.42&78.53&84.27&67.02&68.86&\textbf{77.34}&75.61&59.82&82.01&82.64&67.92&84.19&73.89\\
0.2&59.12&78.83&\textbf{85.12}&66.92&69.04&77.14&75.23&63.58&82.40&82.04&67.81&84.02&74.27\\
0.3&\textbf{60.06}&\textbf{79.05}& 84.48&\textbf{67.49}&\textbf{69.13}&77.03&\textbf{76.03}& \textbf{64.12}&\textbf{82.77}&81.91&\textbf{68.03}&\textbf{84.43}&\textbf{74.55}\\
0.4&60.00&78.64&84.25&67.14&68.92&77.12&75.92&63.78&82.52&81.06&67.98&83.90&74.27\\
0.5&59.43&78.52&83.96&66.82&68.26&76.59&75.98&62.65&82.28&81.42&66.52&84.11&73.88\\
 \bottomrule
 \end{tabular}
 }
 \label{tab:eta}
\end{table*}
\begin{table*}[t]
\centering
\caption{Sensitivity to $\epsilon$ on the {\bf Office-Home} dataset under the {\bf PDA} setting with ResNet-50 as the backbone.}
\vskip -0.1in
\resizebox{\linewidth}{!}{
\begin{tabular}{l|cccccccccccccc}
\toprule
 $\epsilon$ & Ar$\rightarrow$Cl & Ar$\rightarrow$Pr & Ar$\rightarrow$Rw & Cl$\rightarrow$Ar & Cl$\rightarrow$Pr & Cl$\rightarrow$Rw & Pr$\rightarrow$Ar & Pr$\rightarrow$Cl & Pr$\rightarrow$Rw & Rw$\rightarrow$Ar & Rw$\rightarrow$Cl & Rw$\rightarrow$Pr & Avg\\
\midrule
\midrule
0 &60.06&79.05& 84.48&67.49&69.13&77.03&76.03& 64.12&82.77&81.91&68.03&84.43&74.55\\
0.05&\textbf{62.15} & \textbf{84.59} & 85.48 & \textbf{71.17} & \textbf{74.34} &\textbf{80.89}& 78.15 & \textbf{64.72} & \textbf{84.65} & \textbf{82.60} & \textbf{69.43} &\textbf{87.50} & \textbf{77.1}4\\
0.1&61.23&83.44&\textbf{85.74}&70.92&73.44&80.02&\textbf{78.67}&63.78&84.23&81.97&69.02&86.48&76.58\\
0.2& 62.00&82.56&83.22&69.40&72.14&78.58&77.42&63.70&83.28&80.83&68.28&85.62&75.59\\
 \bottomrule
 \end{tabular}
 }
 \label{tab:epsilon}
\end{table*}
% Under the most challenging UniDA settings, according to results shown in Tables \ref{tab:office31-universal}, \ref{tab:officehome-universal}, and \ref{tab:visda}, we can see that the proposed LIWUDA method performs better than state-of-the-art methods in all the datasets. In the challenging Office-Home dataset, the proposed LIWUDA method demonstrates a stronger capability to discriminate between unknown classes and known classes. Specifically, on the Office-31 dataset, the proposed \mbox{LIWUDA} method obtains the best average performance with an improvement of about \textbf{9\%} in terms of the H-score and has the best performance under most tasks. On the Office-Home dataset, the proposed LIWUDA method achieves the best average performance and performs the best on {\bf 8 out of 12} transfer tasks, while it ranks the second or third on the rest 4 transfer tasks. Specifically, compared with the state-of-the-art method CMU, the proposed LIWUDA obtains a significant increase of \textbf{1.16\%} H-score. The proposed method performs a little bit worse in tasks where the target domain is Clipart, and one reason is that the Clipart domain is more complex and contains some noisy data. For the VisDA dataset, the proposed \mbox{LIWUDA} method has the best performance in terms of both evaluation metrics. Specifically, our method gives an improvement of \textbf{2.83\%} accuracy and \textbf{4.05\%} H-score than the SoTA method CMU. 

\subsubsection{Results for PDA}
Based on the results shown in Tables \ref{tab:office31-partial}, \ref{tab:officehome-partial}, and \ref{tab:visda} on the three datasets in the PDA setting, it is discernible that the proposed LIWUDA method surpasses baseline counterparts such as DRCN \cite{LiLLWSHD21}, RTNet \cite{chen2020selective}, and BA$^3$US \cite{liang2020balanced} across diverse datasets. Specifically, the proposed LIWUDA method attains preeminence across all tasks, displaying a nearly {\bf 100\%} accuracy on {\bf 4 out of 6} transfer tasks on the Office-31 dataset. On the Office-Home dataset, the proposed LIWUDA method notably outshines all benchmark methods, yielding the highest average accuracy. Specifically, the proposed LIWUDA enhances the baseline outcome by \textbf{1.16\%}, culminating in an impressive \textbf{77.14\%} accuracy. On the VisDA dataset, the proposed LIWUDA attains the state-of-the-art results, decisively surpassing the most competent baseline method by a notable margin of approximately {\bf 18.53\%} in the S$\rightarrow$R task and an average improvement of {\bf 6.8\%}. Evidently, these findings underscore the effectiveness of our proposed method in effectively bridging domain disparities and elevating classification performance within the target domain under the PDA setting.

\subsubsection{Results for OSDA}

Based on the results shown in Tables \ref{tab:office31-openset}, \ref{tab:officehome-openset}, and \ref{tab:visda} under the OSDA setting, the proposed LIWUDA method yields better performance compared with baseline methods tailed for the OSDA setting. For instance, on the Office-31 and VisDA datasets, the proposed LIWUDA method surpasses OSBP \cite{SaitoYUH18}, STA \cite{LiuCLW019}, and ROS \cite{BucciLT20}, underscoring its enhanced capability in effectively distinguishing between common classes and private classes. While the enhancement in performance compared to ROS might appear marginal, the proposed LIWUDA approach performs classification by adopting an end-to-end approach, facilitating classification across all the four UDA settings, as opposed to the two-stage training paradigm specific to ROS, exclusive to the OSDA framework. Additionally, on the VisDA dataset, the LIWUDA method achieves a noteworthy \textbf{68.4\%} OS$^{*}$ and attains a substantial \textbf{3.9\%} absolute improvement, underscoring its potency in solving OSDA problem.

\begin{figure*}[t]
\centering
\subcaptionbox{ResNet-50}{
\includegraphics[width=0.23\textwidth]{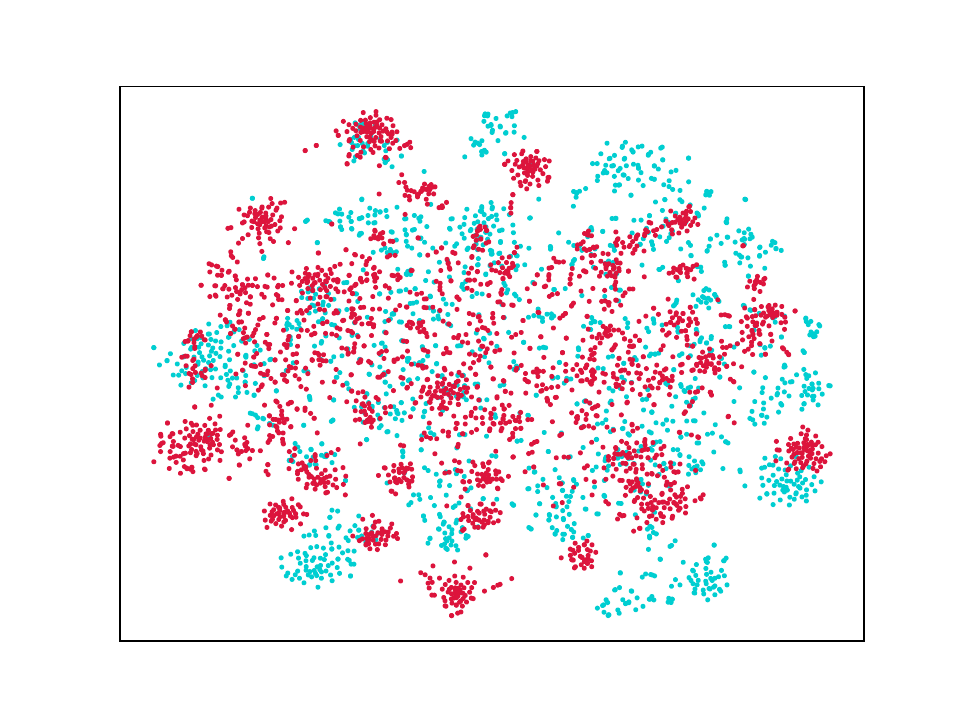}
}
\subcaptionbox{DANN}{
\includegraphics[width=0.23\textwidth]{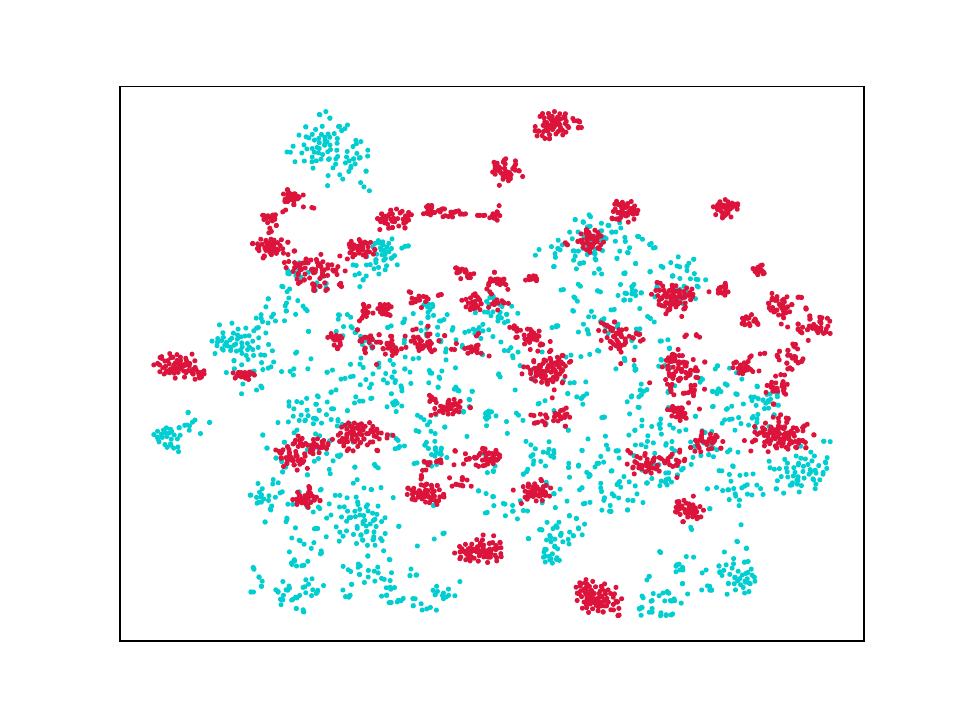}
}
\subcaptionbox{BA$^3$US}{
\includegraphics[width=0.23\textwidth]{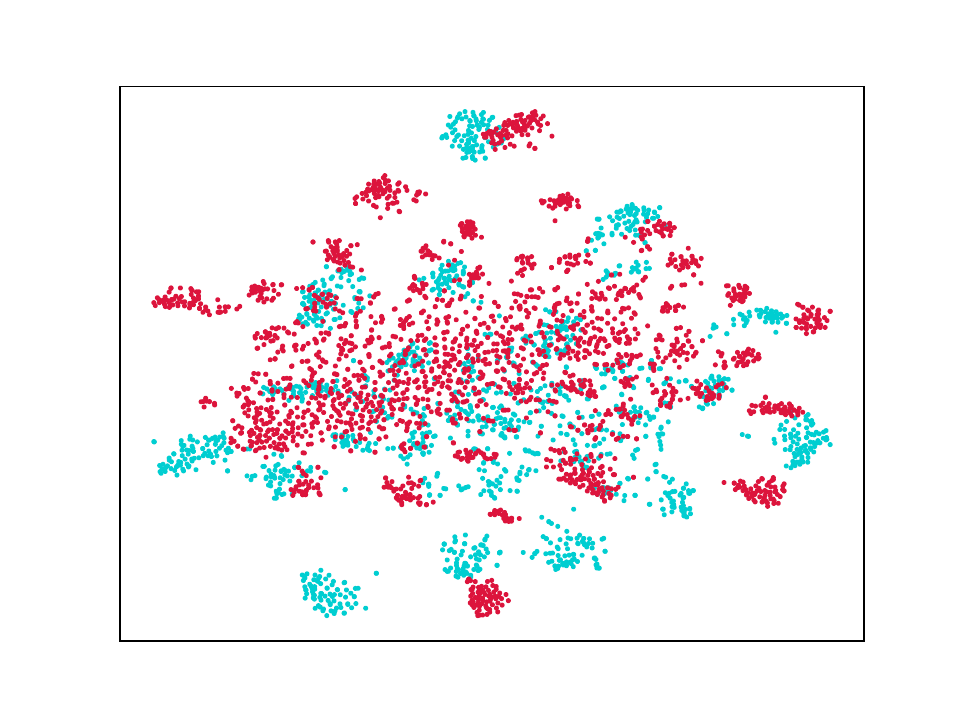}
}
\subcaptionbox{LIWUDA}{
\includegraphics[width=0.23\textwidth]{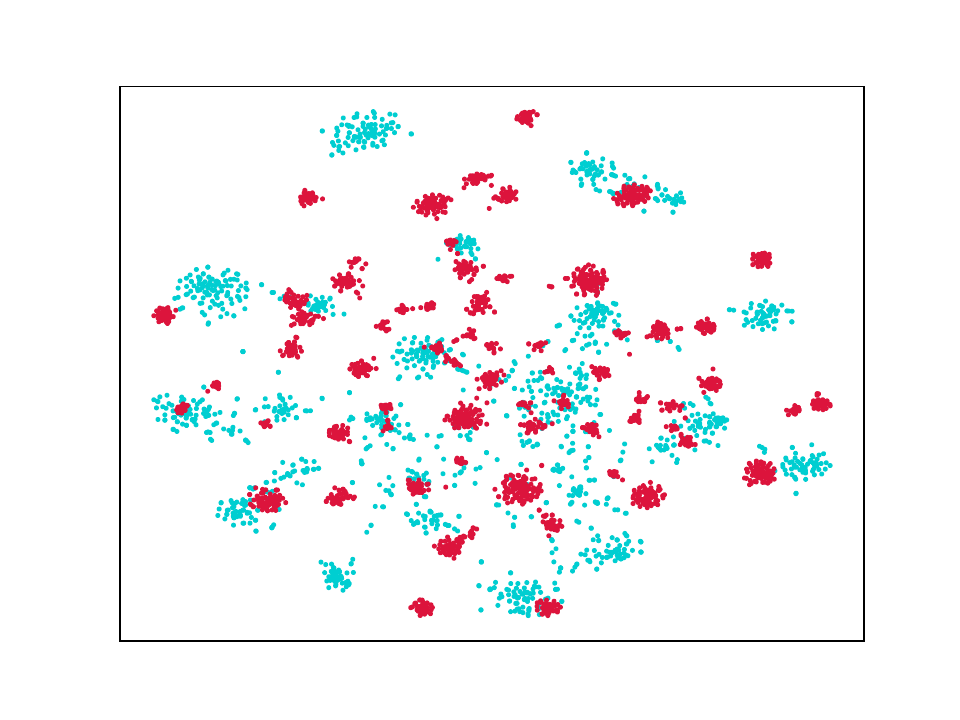}
}
\vskip -0.1in
\caption{t-SNE visualizations for the transfer task Ar$\rightarrow$Cl on the Office-Home dataset under the PDA setting. Source and target instances are shown in red and blue, respectively.}
\label{fig:visualization}
\vskip -0.1in
\end{figure*}

\begin{figure} [t]
\vskip -0.1in
  \centering
    \includegraphics[width=0.49\textwidth]{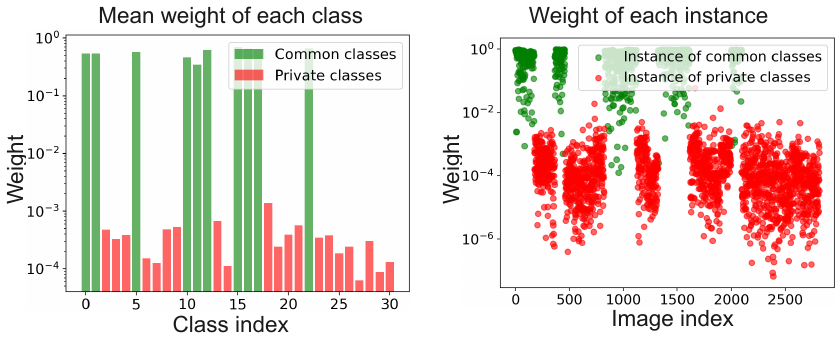}
  \caption{The average weight of each class and weights of instances in a transfer task {\bf A}$\rightarrow$ {\bf W} from the Office-31 dataset under the PDA setting. The green color represents instances from common classes and the red color indicates that instances are from private classes.}
\label{fig:weights}
\end{figure}
% Although compared with ROS, the performance improvement of the proposed LIWUDA method is a bit marginal, the proposed LIWUDA method utilizes an end-to-end approach to make classification under four UDA settings instead of two-stage training of ROS specific to the OSDA setting only. Moreover, on the VisDA dataset, the LIWUDA method achieves \textbf{68.4\%} OS$^{*}$ and leads to \textbf{3.9\%} absolute improvement.
\begin{table*}[t]
\vskip -0.1in
\centering
\vspace{0.2cm}
\caption{Ablation study on \textbf{Office-31} dataset under the PDA setting with ResNet-50 as the backbone.}
\resizebox{\linewidth}{!}{
\begin{tabular}{l|c|c|c||ccccccc}
\toprule
$\mathcal{L}_{\mathcal{C}}$&$\mathcal{L}_{wot}$&$ \mathcal{L}_{sa}$ & $\mathcal{L}_{iot}$& A$\rightarrow$W & D$\rightarrow$W & W$\rightarrow$D & A$\rightarrow$D & D$\rightarrow$A & W$\rightarrow$A & Avg \\
\midrule
\makecell[c]{$\surd$}&&&& 75.59$\pm$1.09 & 96.27$\pm$0.85 & 98.09$\pm$0.74 & 83.44$\pm$1.12 & 83.92$\pm$0.95 & 84.97$\pm$0.86 & 87.05 \\
\makecell[c]{$\surd$}&\makecell[c]{$\surd$}&&&99.52$\pm$ 0.31&99.13$\pm$0.21&{\bf 100.00}$\pm$0.00&$99.42\pm$ 0.13&94.78$\pm$0.25&95.02$\pm$ 0.20&97.98  \\
\makecell[c]{$\surd$}&\makecell[c]{$\surd$}&\makecell[c]{$\surd$}&&99.67$\pm$ 0.26 &{\bf 100.00}$\pm$0.00&{\bf 100.00}$\pm$0.00&{\bf 100.00}$\pm$0.00&95.00$\pm$0.16&95.72$\pm$0.21&98.40\\
\makecell[c]{$\surd$}&\makecell[c]{$\surd$}&\makecell[c]{$\surd$}&\makecell[c]{$\surd$} & {\bf 99.71}$\pm$0.10 & {\bf 100.00}$\pm$0.00 & {\bf 100.00}$\pm$0.00 & 99.42$\pm$0.12 & {\bf 96.33}$\pm$0.15 & {\bf 96.62}$\pm$0.10 & {\bf 98.68}\\
 \bottomrule
 \end{tabular}
 }
%  \vskip 0.1in
 \label{tab:ablation study1}
\end{table*}
\begin{table*}[t]
\vskip -0.1in
\centering
\vspace{0.1cm}
\caption{Ablation study on the {\bf Office-Home} dataset under the PDA setting with ResNet-50 as the backbone.}
\resizebox{\linewidth}{!}{
\begin{tabular}{l|c|c|c||ccccccccccccccc}
\toprule
$\mathcal{L}_{\mathcal{C}}$&$\mathcal{L}_{wot}$&$\mathcal{L}_{sa}$ &$\mathcal{L}_{iot}$& Ar$\rightarrow$Cl & Ar$\rightarrow$Pr & Ar$\rightarrow$Rw & Cl$\rightarrow$Ar & Cl$\rightarrow$Pr & Cl$\rightarrow$Rw & Pr$\rightarrow$Ar & Pr$\rightarrow$Cl & Pr$\rightarrow$Rw & Rw$\rightarrow$Ar & Rw$\rightarrow$Cl & Rw$\rightarrow$Pr & Avg \\
\midrule
\makecell[c]{$\surd$}&&&&46.33 & 67.51 & 75.87 & 59.14 & 59.94 & 62.73 & 58.22 & 41.79 & 74.88 & 67.40 & 48.18 & 74.17 & 61.35\\
\makecell[c]{$\surd$}&\makecell[c]{$\surd$}&&&58.63&79.64&83.10&67.40&68.46&76.42&75.48&57.73&82.33&{\bf 83.10}&65.97&84.09&73.53  \\
\makecell[c]{$\surd$}&\makecell[c]{$\surd$}&\makecell[c]{$\surd$}&&60.06&79.05& 84.48&67.49&69.13&77.03&76.03& 64.12&82.77&81.91&68.03&84.43&74.55\\
\makecell[c]{$\surd$}&\makecell[c]{$\surd$}&\makecell[c]{$\surd$}&\makecell[c]{$\surd$} &{\bf 62.15}$\pm$0.15 & \textbf{84.59}$\pm$0.15 & {\bf85.48} $\pm$0.14 & {\bf71.17}$\pm$0.13 & \textbf{74.34}$\pm$0.13 & {\bf 80.89}$\pm$0.16& {\bf 78.15}$\pm$0.11 & {\bf 64.72}$\pm$0.21 & {\bf 84.65}$\pm$0.24 & 82.60$\pm$0.12 & \textbf{69.43}$\pm$0.23 & {\bf 87.5}$\pm$0.18 & \textbf{77.14}  \\
 \bottomrule
 \end{tabular}
 }
 \label{tab:ablation study2}
\end{table*}

% \subsubsection{Results for CSDA} 
% Performance on the Office-31 and Office-Home datasets is presented in Tables \ref{tab:office31-csda} and \ref{tab:officehome-csda}, respectively. The results underscore that LIWUDA surpasses ResNet-50 \cite{he2016deep}, OTDA \cite{DamodaranKFTC18}, and DeepJDOT \cite{CourtyFTR17}. Notably, the proposed LIWUDA method yields a \textbf{0.7\%} and \textbf{8.2\%} improvement over the respective optimal transport-based baselines, culminating in an accuracy of \textbf{87.5\%} and \textbf{58.9\%} on the two datasets. These findings underscore the effectiveness of the LIWUDA approach in achieving superior domain alignment while preserving class information. In summation, the proposed LIWUDA methodology consistently attains commendable performance across varied UDA settings, highlighting its robust effectiveness in addressing UDA challenges.

% We report the performance on the Office-31 and Office-Home datasets in Tables \ref{tab:office31-csda} and \ref{tab:officehome-csda}, respectively.
% According to the results, LIWUDA outperforms ResNet-50 \cite{he2016deep}, OTDA \cite{DamodaranKFTC18}, and DeepJDOT \cite{CourtyFTR17}. And the proposed LIWUDA improves the baseline result by \textbf{1.3\%} and \textbf{8.2\%}, and yields \textbf{87.5\%} and \textbf{58.9\%} accuracy on the two datasets, respectively. The results demonstrate the LIWUDA method could align domains better with class information. In summary, the proposed LIWUDA method achieves good performance under different UDA settings, which demonstrates its effectiveness on UDA problems.
\subsection{Sensitivity Analysis}
In this section, our experimentation aims to gauge the sensitivity of the performance of the LIWUDA method in relation to its hyperparameters (i.e., $\beta$, $\eta$, and $\epsilon$). The ensuing results are presented in Tables \ref{tab:beta}, \ref{tab:eta}, and \ref{tab:epsilon}, wherein the sets of potential values for these hyperparameters are defined as $\{0, 0.05, 0.1, 0.2, 0.5\}$, $\{0, 0.1, 0.2, 0.3, 0.4, 0.5\}$, and $\{0, 0.05, 0.1, 0.2\}$, correspondingly. The analysis reveals that performance associated with non-zero hyperparameter values consistently outpaces that linked to zero hyperparameter values, thereby underscoring the effectiveness of the proposed loss functions embedded within the LIWUDA method. Additionally, the pinnacle of performance materializes when $\beta$, $\eta$, and $\epsilon$ are respectively set to 0.1, 0.3, and 0.05. Consequently, these settings are embraced across all experiments to ensure optimal performance across all UDA scenarios.

% In this section, we conduct experiments to test the sensitivity of the performance of the LIWUDA method with respect to hyperparameters (i.e., $\beta$, $\eta$, and $\epsilon$) and show the results in Tables \ref{tab:beta}, \ref{tab:eta}, and \ref{tab:epsilon}, where the sets of candidates for the three hyperparameters are $\{0, 0.05, 0.1, 0.2, 0.5\}$, $\{ 0, 0.1, 0.2, 0.3, 0.4, 0.5 \}$, and $\{0, 0.05, 0.1, 0.2 \}$. According to the results, we can see that the performance corresponding a non-zero hyperparameter is better than that with a zero hyperparameter, which demonstrates the usefulness of the proposed losses in the LIWUDA method. Moreover, the performance achieves the best when $\beta$, $\eta$, and $\epsilon$ takes the value of 0.1, 0.3, and 0.05, respectively, and we will adopt such settings for the three hyperparameters in all the experiments.
\subsection{Analysis on Weights of Instances}

To substantiate the effectiveness of instance weighting within the proposed LIWUDA method, we undertake an experiment on the transfer task {\bf A}$\rightarrow$ {\bf W} using the Office-31 dataset within the PDA setting. The obtained results are visualized in Figure \ref{fig:weights}, illustrating the weights assigned to individual samples as well as the mean weights attributed to samples belonging to the same class within the source domain. From the insights gleaned in Figure \ref{fig:weights},  it can be observed the weights acquired through the proposed LIWUDA method aptly segregate instances from both common and private classes.  This demonstrates the effectiveness of the weight network embedded within the LIWUDA framework, substantiating its role in effectively distinguishing common and private classes.

% To show the effectiveness of instance weighting in the proposed LIWUDA method, we conduct an experiment on transfer task {\bf A}$\rightarrow$ {\bf W} from the Office-31 dataset under the PDA setting, and present in Figure \ref{fig:weights} weights of samples and mean weights of samples within the same class in the source domain. According to Figure \ref{fig:weights}, we can see that weights learned in the proposed LIWUDA method can well separate instances from common and private classes, which demonstrates the usefulness of the weight network in the LIWUDA method.

\subsection{Feature Visualization}
The visualization presented in Figure \ref{fig:visualization} portrays the feature representations acquired by ResNet-50 \cite{he2016deep}, DANN \cite{ganin2016domain}, BA$^3$US \cite{liang2020balanced}, and LIWUDA on the transfer task Ar $\rightarrow$ Cl from the Office-Home dataset under the PDA setting. This visualization is facilitated through the t-SNE method \cite{DonahueJVHZTD14}. Comparative analysis against baseline methods reveals a notable prowess of the proposed LIWUDA method. It demonstrates an impressive ability to align data from the same class across diverse domains, while concurrently separating data from distinct classes. This result serves as a visual testament to the capability of the LIWUDA model to adeptly align two domains under disparate label spaces in this task. This alignment factor, to some extent, contributes to the noteworthy performance exhibited by the LIWUDA method.

\subsection{Ablation Study}
To evaluate the individual contributions of $\mathcal{L}_{wot}$, $\mathcal{L}_{sa}$, and $\mathcal{L}_{iot}$, respectively, we train the LIWUDA model with various combinations of these losses under the PDA setting. The results are presented on Office-31 and Office-Home datasets, as detailed in Tables \ref{tab:ablation study1} and \ref{tab:ablation study2}. According to the results, even without the SA loss $\mathcal{L}_{sa}$ and IOT loss $\mathcal{L}_{iot}$, the WOT loss $\mathcal{L}_{wot}$ still achieves impressive results of 97.98\% and 73.53\% on the Office-31 and Office-Home datasets, showcasing the effectiveness of the WOT $\mathcal{L}_{wot}$. Furthermore, the inclusion of the SA loss ($\mathcal{L}_{sa}$) leads to a notable improvement in classification performance by \textbf{0.42}\% and \textbf{1.02}\% on the Office-31 and Office-Home datasets, affirming its effectiveness. When comparing models with or without the IOT loss ($\mathcal{L}_{iot}$), our method demonstrates a substantial increase by \textbf{0.28}\% and \textbf{2.59}\% separately. These results underscore the significant contribution of the SA loss ($\mathcal{L}_{sa}$) in separating/aligning samples, while the IOT loss ($\mathcal{L}_{iot}$) aids in achieving similar weights for common classes. 
In summary, the comprehensive LIWUDA method demonstrates a substantial performance boost, underscoring the synergistic effectiveness of the integrated three losses.

% Moreover, the SA loss $\mathcal{L}_{sa}$ boosts the improvement of classification to \textbf{0.42}\% and \textbf{1.02}\% on the Office-31 and Office-home datasets, which verifies their effectiveness. 

% Compared with the model with or without IOT loss $\mathcal{L}_{iot}$, our method achieves a dramatic increase by \textbf{0.28}\% and \textbf{2.59}\% separately. The results demonstrate the SA loss $\mathcal{L}_{sa}$ has the great ability to separate/align samples and the IOT loss $\mathcal{L}_{iot}$ could help to make the weight distributions of common classes similar. In summary, the whole LIWUDA method performs the best in almost all the transfer tasks, which implies that the combination of the four losses works very well.
%that SA has the great ability to separate$/$ align samples and IOT could help to make the weight distributions of common classes similar.

\section{Conclusion}

In this work, we propose the LIWUDA method, a unified framework tailored for four distinct UDA settings accomplished through the strategic application of instance weighting. Specifically, we introduce the WOT to mitigate the domain divergence by alleviating the negative transfer caused by domain-specific classes. Additionally, we introduce the SA loss to effectively separate instances with low similarities and align instances that exhibit high similarities. To enhance the learning process of the weight network, we incorporate IOT, thereby providing supplementary insights. The integration of these three components results in the comprehensive LIWUDA framework for all four UDA settings and yields competitive performance across three benchmark datasets.

While the proposed LIWUDA method is specifically designed for UDA settings, its applicability is limited to this domain. In future work, we aim to extend the LIWUDA method to handle other settings in domain adaptation and transfer learning, such as semi-supervised domain adaptation, multi-source domain adaptation, and source-free domain adaptation.

\section*{Acknowledgements}

This work is supported by NSFC key grant 62136005, NSFC general grant 62076118, and Shenzhen fundamental research program JCYJ20210324105000003.

%%%%%%%%% REFERENCES
\bibliographystyle{IEEEtran}
\bibliography{IEEEabrv, egbib}

\vfill

\end{document}